\documentclass[sigconf, nonacm]{acmart}

\newcommand\vldbdoi{10.14778/3696435.3696439}
\newcommand\vldbpages{42 - 52}
\newcommand\vldbvolume{18}
\newcommand\vldbissue{1}
\newcommand\vldbyear{2024}
\newcommand\vldbauthors{\authors}
\newcommand\vldbtitle{\shorttitle} 
\newcommand\vldbpagestyle{empty} 
\newcommand\vldbavailabilityurl{https://github.com/fpgasystems/Chameleon-RAG-Acceleration}

\newcommand{\red}[1]{#1}

\usepackage{tikz}
\usepackage{amsmath}

\usepackage{fontawesome}
\usepackage{enumitem}
\usepackage{hyperref}
\usepackage{multicol}
\usepackage{multirow}
\usepackage{tablefootnote}
\usepackage{soul}
\usepackage{calrsfs}
\usepackage{xcolor}

\usepackage[super]{nth}

\usepackage{tikz}
\usepackage{xcolor}

\newcommand{\ballnumber}[1]{\tikz[baseline=(myanchor.base)] \node[circle,fill=.,inner sep=1pt] (myanchor) {\color{-.}\bfseries\footnotesize #1};}
\newcommand{\blueballnumber}[1]{\tikz[baseline=(myanchor.base)] \node[circle,fill=blue!80,inner sep=1pt] (myanchor) {\color{-.}\bfseries\footnotesize #1};}

\newcommand{\whiteballnumber}[1]{\tikz[baseline=(myanchor.base)] \node[circle,fill=black!0,inner sep=1pt, draw=black!100, thick] (myanchor) {\color{black}\bfseries\footnotesize #1};}

\usepackage{tabularx}
\newcolumntype{M}[1]{>{\centering\arraybackslash}m{#1}} 
\newcolumntype{R}[1]{>{\raggedleft\arraybackslash}m{#1}} 
\newcolumntype{L}[1]{>{\raggedright\arraybackslash}m{#1}} 

\usepackage{booktabs}

\usepackage{arydshln}

\usepackage{wrapfig}
\usepackage{graphicx}
\usepackage{subcaption}
\usepackage{amsmath}
\usepackage[ruled,vlined]{algorithm2e}

\usepackage[most]{tcolorbox}

\usepackage[normalem]{ulem}

\begin{document}

\title{Chameleon: a Heterogeneous and Disaggregated Accelerator System for Retrieval-Augmented Language Models}

\date{}




\author{Wenqi Jiang}
\affiliation{%
  \country{Systems Group, ETH Zurich}
}
\email{wenqi.jiang@inf.ethz.ch}

\author{Marco Zeller}
\affiliation{%
  \country{Systems Group, ETH Zurich}
}
\email{mzeller@student.ethz.ch}

\author{Roger Waleffe}
\affiliation{%
  \country{University of Wisconsin Madison}
}
\email{waleffe@wisc.edu}

\author{Torsten Hoefler}
\affiliation{%
  \country{SPCL, ETH Zurich}
}
\email{torsten.hoefler@inf.ethz.ch}

\author{Gustavo Alonso}
\affiliation{%
  \country{Systems Group, ETH Zurich}
}
\email{alonso@inf.ethz.ch}

\sloppy
\begin{abstract}

A Retrieval-Augmented Language Model (RALM) combines a large language model (LLM) with a vector database to retrieve context-specific knowledge during text generation.
This strategy facilitates impressive generation quality even with smaller models, thus reducing computational demands by orders of magnitude.
To serve RALMs efficiently and flexibly, we propose \textit{Chameleon}, a heterogeneous accelerator system integrating both LLM and vector search accelerators in a disaggregated architecture. 
The \textit{heterogeneity} ensures efficient serving for both inference and retrieval, while the \textit{disaggregation} allows independent scaling of LLM and vector search accelerators to fulfill diverse RALM requirements. 
Our Chameleon prototype implements vector search accelerators on FPGAs and assigns LLM inference to GPUs, with CPUs as cluster coordinators.
Evaluated on various RALMs, Chameleon exhibits up to 2.16$\times$ reduction in latency and 3.18$\times$ speedup in throughput compared to the hybrid CPU-GPU architecture.
The promising results pave the way for adopting heterogeneous accelerators for not only LLM inference but also vector search in future RALM systems.



\end{abstract}

\maketitle 

\pagestyle{\vldbpagestyle}
\begingroup\small\noindent\raggedright\textbf{PVLDB Reference Format:}\\
\vldbauthors. \vldbtitle. PVLDB, \vldbvolume(\vldbissue): \vldbpages, \vldbyear.\\
\href{https://doi.org/\vldbdoi}{doi:\vldbdoi}
\endgroup
\begingroup
\renewcommand\thefootnote{}\footnote{\noindent
This work is licensed under the Creative Commons BY-NC-ND 4.0 International License. Visit \url{https://creativecommons.org/licenses/by-nc-nd/4.0/} to view a copy of this license. For any use beyond those covered by this license, obtain permission by emailing \href{mailto:info@vldb.org}{info@vldb.org}. Copyright is held by the owner/author(s). Publication rights licensed to the VLDB Endowment. \\
\raggedright Proceedings of the VLDB Endowment, Vol. \vldbvolume, No. \vldbissue\ %
ISSN 2150-8097. \\
\href{https://doi.org/\vldbdoi}{doi:\vldbdoi} \\
}\addtocounter{footnote}{-1}\endgroup

\ifdefempty{\vldbavailabilityurl}{}{
\vspace{.3cm}
\begingroup\small\noindent\raggedright\textbf{PVLDB Artifact Availability:}\\
The source code, data, and/or other artifacts have been made available at \url{\vldbavailabilityurl}.
\endgroup
}

\section{Introduction}
\label{sec:intro}

\begin{figure}[t]
	\centering
  \includegraphics[width=0.95\linewidth]{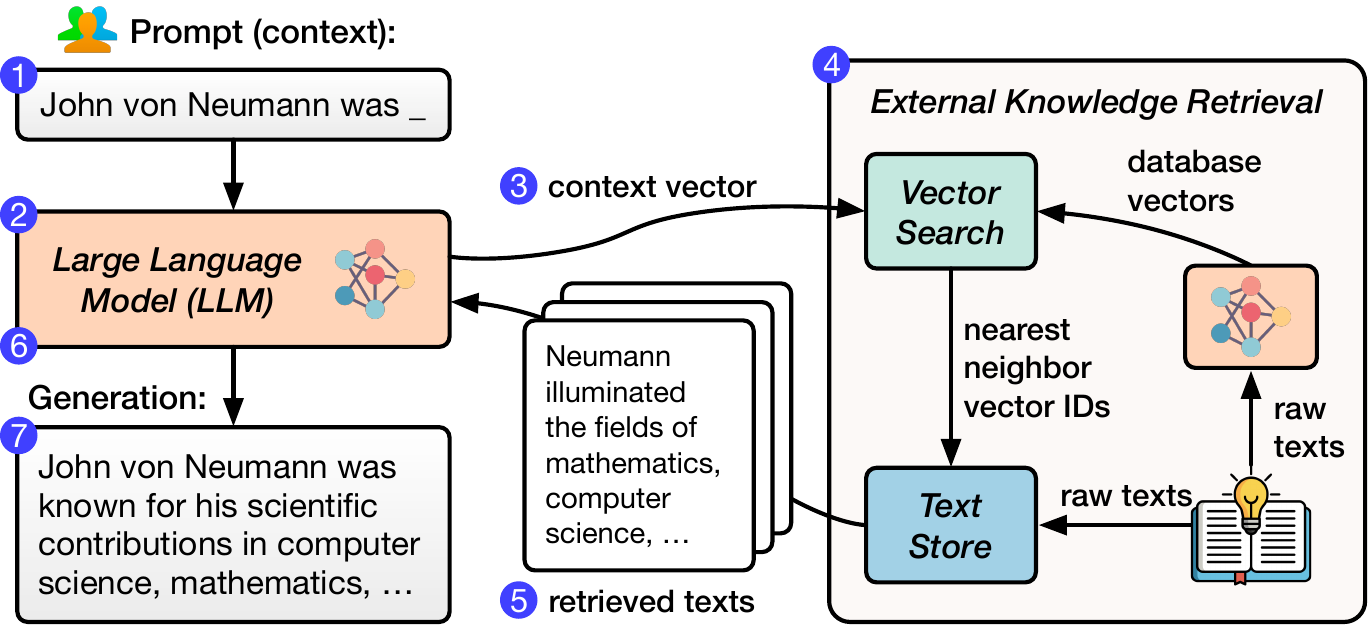}
  \vspace{-1em}
  \caption{A retrieval-augmented language model (RALM).}
  \vspace{-2em}
  \label{fig:ralm}
\end{figure}

Vector databases facilitate the development of \textit{Retrieval-Augmented Language Models (RALMs)}, an increasingly popular approach to serve generative large language models (LLMs).
The architecture of RALM, as shown in Figure~\ref{fig:ralm}, allows the LLM to focus on learning linguistic structures, while incorporating context-specific knowledge during inference. 
Specifically, the external textual knowledge is encoded as vectors using LLMs and stored in a vector database. 
Given an inference context (e.g., a prompt), the knowledge retriever identifies relevant knowledge in the database via vector search, which assesses relevance by computing the similarity between the context vector and the database vectors. The retrieved texts are then incorporated into the LLM to facilitate high-quality generation. 

RALMs show three major advantages over conventional LLMs.
\textit{First of all}, RALMs, even using smaller LLMs with one to two orders of magnitude fewer parameters, can match or surpass the generation quality of conventional LLMs on various tasks~\cite{lewis2020retrieval, izacard2020leveraging, komeili2021internet, guu2020retrieval, khandelwal2019generalization, khandelwal2020nearest, lewis2020pre}, thus significantly lowering the inference cost. 
This is because conventional LLMs rely on a vast number of parameters trained on massive datasets to capture and retain textual knowledge~\cite{brown2020language, chowdhery2022palm, smith2022using, rae2021scaling}, while RALMs can integrate retrieved knowledge during inference, not burdening the LLM's parameters.
\textit{Moreover}, knowledge editing in RALMs is as straightforward as updating the database, enabling efficient integration of new or private knowledge~\cite{OpenAI_assistant_2, OpenAI_assistant_1}. In contrast, updating knowledge in conventional LLMs is inflexible, requiring additional training~\cite{borgeaud2022improving, lewis2020retrieval}. 
\textit{Finally}, RALMs enhance the reliability and interpretability of generated content by sourcing knowledge externally, while conventional LLMs are prone to producing non-factual content, known as hallucination~\cite{lewis2020retrieval, li2023dark}.

Despite its advantages, efficient RALM inference presents \textbf{two challenges}. \textit{First, the workload characteristics of the LLM and the retriever are distinct.} While the LLM inference primarily relies on rapid tensor operations, the vector search system --- often utilizing fast and memory-efficient search algorithms like Product Quantization (PQ)~\cite{PQ} --- demands both substantial memory capacity to hold the vectors and fast processing of quantized database vectors during query time.
\textit{Second, the diverse range of RALM configurations leads to shifting system requirements and bottlenecks.}
\red{Regarding retrieval frequency, some models retrieve once per generated token~\cite{khandelwal2019generalization, meng2021fast, alon2022neuro}, while others retrieve only once per entire sequence~\cite{lewis2020retrieval, izacard2020leveraging}. 
In terms of scale, database sizes vary from millions~\cite{lewis2020retrieval, guu2020realm} to tens of billions of vectors (92 TB)~\cite{borgeaud2022improving, wang2023instructretro}, and model sizes range from hundreds of millions~\cite{guu2020realm, lewis2020pre} to tens of billions of parameters~\cite{wang2023instructretro}.}


We envision a high-performance and efficient RALM system to adhere to \textbf{two key design principles} to address the two aforementioned challenges. 
\textit{Firstly, RALMs should incorporate \textbf{heterogeneous accelerators}, employing not only inference accelerators such as GPUs but also vector search accelerators}, such that both RALM components are fast and efficient.
\textit{Secondly, the heterogeneous accelerators should be \textbf{disaggregated} to support diverse RALM demands efficiently}, in contrast to a monolithic approach where a fixed number of LLM and retrieval accelerators reside on the same server. 
The rationale is twofold: (a) performance bottlenecks shift between various RALMs of different retrieval frequencies, database sizes, and model sizes, thus requiring a case-specific optimal balance between the two types of accelerators; and (b) \red{a huge database (e.g., with tens of TBs of vectors~\cite{borgeaud2022improving, wang2023instructretro}) may necessitate more retrieval accelerators than a single server can accommodate.}



To materialize this vision, we propose \textit{Chameleon}, a heterogeneous and disaggregated accelerator system for efficient, flexible, and high-performance RALM inference.
Chameleon consists of three primary components.
Firstly, \textit{ChamVS} is a distributed and accelerated vector search engine. It consists of several disaggregated memory nodes, each containing a shard of quantized database vectors in DRAM, a near-memory retrieval accelerator prototyped on FPGA, and a hardware TCP/IP stack. 
Secondly, \textit{ChamLM} is a multi-GPU LLM inference engine. It produces query vectors and generates texts using the retrieved information.
Lastly, a CPU coordinator server orchestrates the network communication between the retrieval and LLM accelerators.

We evaluate Chameleon with various LLM architectures, model sizes, database sizes, and retrieval frequencies. 
For large-scale vector search, ChamVS achieves up to 23.72$\times$ latency reduction compared to the optimized CPU baselines while consuming 5.8$\sim$26.2$\times$ less energy. 
For RALM inference, Chameleon achieves up to 2.16$\times$ and 3.18$\times$ speedup in latency and throughput compared to the hybrid CPU-GPU architecture. 
We further illustrate that the optimal balance between the two types of accelerators varies significantly across different RALMs, making disaggregation essential for achieving both flexibility and high accelerator utilization rates.



The paper makes the following \textbf{contributions:}

\begin{itemize}[leftmargin=*]
    \item We present Chameleon, an efficient RALM inference system designed around two proposed principles: accelerator heterogeneity and disaggregation. 
    \item We design and implement ChamVS, a distributed engine for large-scale vector search, which includes:
    \begin{itemize}
	\item Near-memory accelerators for vector search, including a novel resource-efficient top-K selection architecture.
        \item A GPU-based index scanner to prune search space.
    \end{itemize}
    \item We evaluate Chameleon on various RALMs and showcase its remarkable performance and efficiency.
\end{itemize}

\section{Background and Motivation}
\label{sec:background}


\subsection{Retrieval-Augmented Language Models}
\label{sec:background_ralm}

A RALM combines an LLM~\cite{radford2019language, devlin2018bert, raffel2020exploring} with a vector database. During inference, information relevant to the current context is retrieved from the database and utilized by the LLM to predict subsequent tokens. 
\textit{We classify RALMs by the content they retrieve:}

The first category of RALMs retrieves \textit{text chunks} containing multiple tokens related to the current context~\cite{borgeaud2022improving, lewis2020retrieval, izacard2020leveraging, ram2023context, jiang2024piperag}. 
During inference, the generation context, such as a user's prompt, is encoded as a query vector to retrieve context-related knowledge, i.e., text chunks in the database with similar vector representations~\cite{borgeaud2022improving, lewis2020retrieval, izacard2022few}. 
The retrieved text chunks are then integrated by the LLM, leading to the generation of output tokens. 
When generating long sequences, however, the generated content may gradually diverge from the initially retrieved contents. Thus, instead of initiating retrieval only once at the beginning~\cite{lewis2020retrieval, izacard2020leveraging, sachan2021end}, an effective strategy is to perform multiple retrievals during text generation to improve token generation quality~\cite{ram2023context}, for instance, at a regular interval of every 8$\sim$64 generated tokens~\cite{borgeaud2022improving, ram2023context, jiang2024piperag}.

The second category of RALMs retrieves only the \textit{next token} of each similar context in the database~\cite{khandelwal2019generalization, meng2021fast, alon2022neuro}. 
At each step of token generation (retrieval interval is one), the last layer's hidden state serves as the query to retrieve similar contexts and the next token of each similar context~\cite{khandelwal2019generalization, meng2021fast, xu2023nearest}. The next token of the current context is then predicted by interpolating the next-token probability distribution predicted by the model with that of the retrieved content~\cite{khandelwal2019generalization, khandelwal2020nearest}.


\subsection{Large-Scale Vector Search}
\label{sec:background_vs}


A vector search takes a $D$-dimensional query vector $x$  as input and retrieves $K$ similar vector(s) from a database $Y$, populated with many $D$-dimensional vectors, based on metrics like L2 distances or cosine similarity. 
Exact $K$ nearest neighbor (KNN) search can be prohibitively expensive for large datasets, requiring a linear scan through all database vectors.
Thus, real-world vector search systems adopt approximate nearest neighbor (ANN) search that can achieve much higher system performance. The quality of an ANN search is measured by the recall at $K$ ($R@K$), which denotes the overlap percentage between the exact $K$ nearest neighbors and the $K$ returned by the ANN. In the subsequent sections, we will use the terms \textit{vector search} and \textit{ANN search} interchangeably.



\textit{IVF-PQ}, combining the inverted-file (IVF) index and product quantization (PQ), is among the most popular vector search algorithms in RALMs~\cite{lewis2020retrieval, borgeaud2022improving, izacard2020leveraging, khandelwal2019generalization}.
\red{Its more frequent adoption over other ANN algorithms like graph-based vector search~\cite{malkov2014approximate, malkov2018efficient, fu2017fast, zhao2023towards, zuo2023arkgraph, lu2021hvs, gao2023high} is primarily due to memory efficiency: RALMs can involve large databases, with reported sizes reaching up to 30 billion vectors (92 TB)~\cite{borgeaud2022improving, wang2023instructretro}, thus the high compression ratio offered by PQ~\cite{PQ, OPQ, johnson2019billion} is essential.}



\textbf{Inverted-File (IVF) Index.} An IVF index divides a vector dataset $Y$ into many (\textit{nlist}) disjoint subsets, typically using clustering algorithms like K-means. Each of these subsets is termed an IVF list. At query time, the IVF index is scanned, and only a select few (\textit{nprobe}) IVF lists whose cluster centroids are close to the query vector are scanned, such that the search space is effectively pruned. 

\begin{figure}[t]
	\centering
  \includegraphics[width=0.8\linewidth]{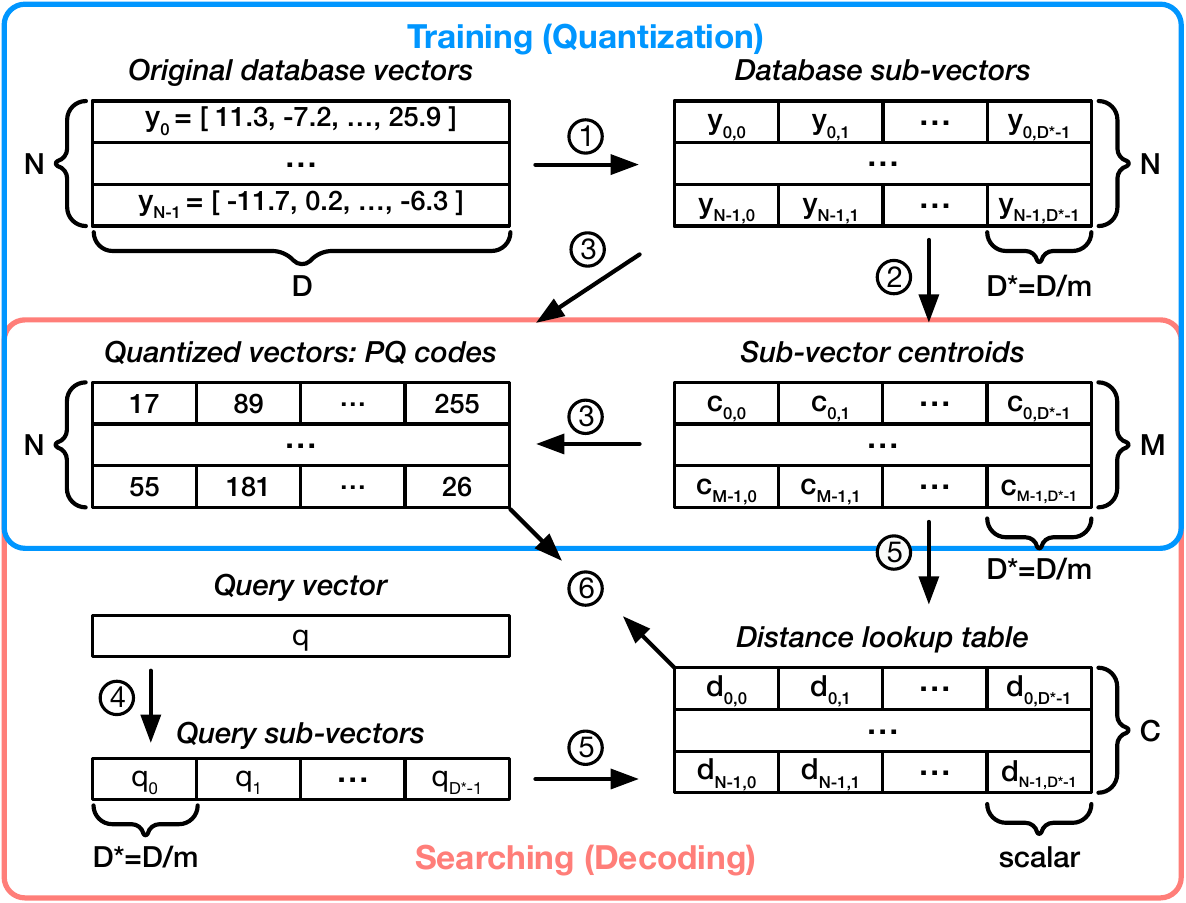}
	\vspace{-1em} 
  \caption{Product quantization (PQ) for vector search.}
  \label{fig:PQ}
	\vspace{-1.5em} 
\end{figure}

\textbf{Product Quantization (PQ).}
PQ reduces memory usage and computations of vector search by compressing each database vector into $m$-byte PQ codes. Figure~\ref{fig:PQ} overviews the workflow of PQ.

\textit{Training (quantization).} All database vectors are partitioned evenly into $m$ sub-vectors~\whiteballnumber{1}, which possess a dimensionality of $D^{*}=\frac{D}{m}$, typically ranging from 4 to 16 in practice. 
A clustering algorithm is performed in each sub-space~\whiteballnumber{2} to obtain a list of centroids $c$, allowing each database sub-vector to be approximated by its nearest centroid. 
Typically, the number of clusters per sub-space is set as $M=256$, such that a cluster ID can be represented with one byte. Thus, once the cluster centroids are stored, each database vector can be represented by $m$-byte PQ codes.


\textit{Searching (decoding).} A query vector is compared against the quantized database vectors.
The distance computation can be formulated as $\hat{d}(x,y)=d(x,c(y)) =\sum_{i=1}^{m}d(x_i,c_i(y_i))$, where $\hat{d}(x,y)$ is the approximate distance between a query vector $x$ and a quantized database vector $y$, and $c(y)$ is the reconstructed database vector using the PQ codes and the cluster centroid vectors per sub-space. 
To calculate $\hat{d}(x,y)$, the query vector is divided into $m$ sub-vectors ($x_i$)~\whiteballnumber{4} and compared against the reconstructed quantized sub-database-vectors $c_i(y_i)$.
To speed up distance computations given many database vectors, a distance lookup table~\whiteballnumber{5} can be constructed and reused within a query, encompassing all combinations between a sub-query-vector and a cluster centroid within the same sub-space. 
With this table, the value of $d(x_i,c_i(y_i))$ can be swiftly retrieved by looking up the table with the PQ code as the address~\whiteballnumber{6}, leading to improved computational efficiency.



\subsection{Motivation: Efficient RALM Inference}
\label{sec:motivation}


An efficient RALM inference engine should meet the following \textbf{system requirements}:

\begin{itemize}[leftmargin=*]
    \item Both the LLM inference and the large-scale vector search components should be fast and resource-efficient. 
    \item The system should be flexible enough to accommodate diverse RALM configurations, spanning various combinations model sizes, database sizes, and retrieval frequencies.
\end{itemize}

However, little effort has been devoted to developing efficient RALM systems that meet the above requirements. This is likely because RALM has been an emerging topic within the machine learning community~\cite{borgeaud2022improving, khandelwal2019generalization, lewis2020retrieval, izacard2022few, izacard2020leveraging}, with their prototype implementations exhibiting the following problems:

\underline{\textbf{(P1)}} Each research RALM system focuses on \textit{being able to run} one or a small number of RALM models, paying little attention to latency, throughput, resource efficiency, and system flexibility.

\underline{\textbf{(P2)}} While hardware accelerators for LLMs, such as GPUs, are advancing rapidly, less attention has been paid to the vector search aspect, which, as our evaluations will demonstrate, can become the performance bottleneck in RALM inference. 

\underline{\textbf{(P2.1)}} CPUs are slow in scanning PQ codes during query time~\whiteballnumber{6}. This is due to the frequent cache accesses (for each byte of PQ code, load the code and use it as an address to load a distance) and the instruction dependencies between operations  (distance lookups depend on PQ codes and distance accumulations depend on the lookup values). 
Even with the state-of-the-art SIMD-optimized CPU implementation~\cite{faiss}, the throughput peaks at roughly 1 GB/s per core when scanning PQ codes (1.2 GB/s on Intel Xeon Platinum 8259CL @ 2.50GHz).
Within a CPU-memory-balanced server, the PQ code scanning process significantly underutilizes the available memory bandwidth, as about 16 cores are required to saturate the bandwidth of a single memory channel (around 20 GB/s).

\underline{\textbf{(P2.2)}} GPUs suffer from two major limitations for large-scale vector search. 
Firstly, the limited memory capacity of each GPU makes large-scale searches on GPU clusters cost-prohibitive. 
For instance, accommodating only 1 TB of PQ codes necessitates at least 16 NVIDIA A100 GPUs (cost 300K USD as of March 2024), each with 80 GB of memory, given that a portion of memory should be reserved for intermediate search states. 
Although an alternative solution is to adopt a hybrid CPU-GPU architecture where the GPU fetches vectors from CPU's memory, the inter-processor bandwidth is way lower than the GPU memory bandwidth. Even for NVIDIA Grace Hopper, with the latest high-performance CPU-GPU interconnect, the single-direction bandwidth of 450 GB/s is only 15\% of the GPU's bandwidth. 
Secondly, the throughput for PQ code scanning on GPUs is considerably lower than the GPU's bandwidth, only around 50\% of the bandwidth even with large batch sizes (evaluated on NVIDIA A100), due to the multiple passes of memory accesses to write and read intermediate results at each search step~\cite{johnson2019billion}. 

\section{Chameleon: System Overview}

\begin{figure*}[t]
	\centering
  \includegraphics[width=0.9\linewidth]{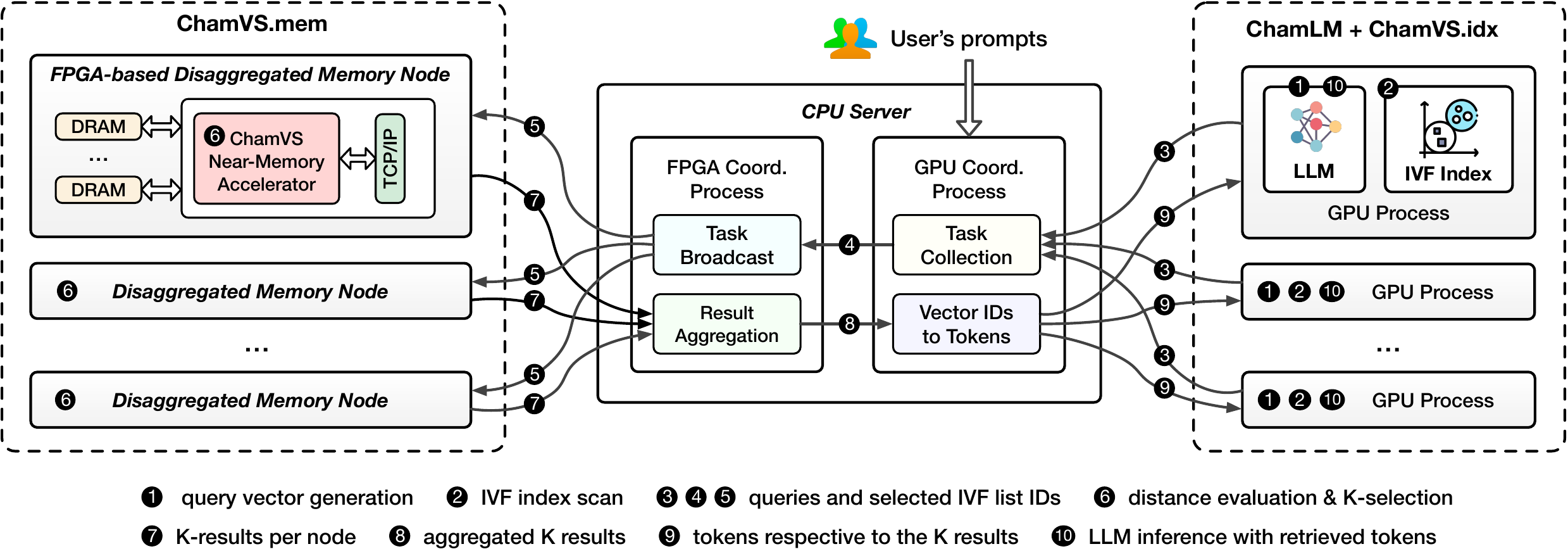}
  \vspace{-1em}
  \caption{Chameleon is a heterogeneous and disaggregated accelerator system for efficient RALM inference.}
  \vspace{-1em}
  \label{fig:overview}
\end{figure*}

\textbf{We design and implement Chameleon, an efficient, flexible, and performant RALM inference system:}



    
\begin{itemize}[leftmargin=*]
    \item Chameleon employs heterogeneous hardware to accelerate both LLM inference and vector search efficiently.
    \item Chameleon disaggregates the accelerators, enabling independent scaling for each type of hardware, thus supporting various RALM configurations efficiently.
    \item The modular design of Chameleon allows flexible hardware upgrades, such as integrating more powerful LLM inference accelerators or ASIC-based ChamVS accelerators in the future.
\end{itemize}

\textbf{Figure~\ref{fig:overview} overviews the Chameleon architecture}, which primarily consists of the following components.

\textit{Firstly, ChamVS is a distributed accelerator engine for low-latency vector search.}
On the one hand, ChamVS.idx is a GPU-based IVF index scanner colocated with the ChamLM GPUs (right side of Figure~\ref{fig:overview}). \red{While Chameleon also supports index scan on CPUs}, GPUs are generally more favorable for handling this embarrassingly parallel workload due to their superior memory bandwidth and computational capability.
Given that GPUs are already integrated into Chameleon, no additional devices are required. 
\red{The only overhead is a slight increase in GPU memory usage, as the index sizes are small relative to the database vectors. For example, assuming 1KB per vector and one thousand vectors per IVF list, a single GB of index can support one million IVF lists, enough for a large database containing one billion vectors.}
On the other hand, ChamVS.mem is responsible for querying quantized database vectors. ChamVS.mem contains one or multiple disaggregated memory nodes, each with a partition of the database vectors and a near-memory retrieval accelerator prototyped on FPGA for query processing (left side of Figure~\ref{fig:overview}). 


\textit{Secondly, ChamLM is a multi-GPU LLM inference engine}, as shown on the right side of Figure~\ref{fig:overview}. 
Each GPU, managed by an independent GPU process, can reside on the same or different servers. 
Currently, ChamLM assigns each GPU a full copy of the LLM, as RALMs can achieve high generation quality even with smaller LLMs~\cite{borgeaud2022improving, lewis2020retrieval}. Future larger models could be accommodated by extending ChamLM to support tensor or pipeline parallelism~\cite{narayanan2019pipedream, shoeybi2019megatron, rajbhandari2020zero} across GPUs.
\red{Once a retrieval request is sent, a GPU pauses inference to wait for results. While one potential solution to avoid such GPU idleness is to split the generation into two sub-batches --- one executes inference when the other one is waiting for retrieved contents --- this approach does not necessarily improve performance. This is because (a) using sub-batches reduces inference throughput, and (b) retrieval latency may not align with inference latency.
 }

\textit{Thirdly, the CPU serves as the cluster coordinator, managing the lightweight communication between the GPUs and FPGAs.} After receiving search requests from the GPU processes, it dispatches them to the FPGA-based disaggregated memory nodes, aggregates the per-partition results returned by the FPGAs, converts the K nearest neighbor vector IDs into their corresponding texts, and sends the retrieved tokens back to the GPUs. Since each query only requires less than ten KBs of network data transfer, the communication latency is negligible compared to vector search and LLM inference. 

\textbf{Token generation workflow.}
For each token generation step, the procedure diverges depending on whether the retrieval is invoked. 
Without retrieval, the GPUs infer the next token as in regular LLMs.
With retrieval, the first step is to generate a contextual query vector~\ballnumber{1}, either by using the hidden state of the current context~\cite{khandelwal2019generalization, khandelwal2020nearest} or encoding the query tokens through another model~\cite{borgeaud2022improving}. 
Following this, the IVF index residing on the same GPU is scanned to select the $nprobe$ most relevant IVF lists~\ballnumber{2}.
The query vector and the list IDs are then transmitted to the GPU coordinator process running on the CPU node via the network~\ballnumber{3}. After recording the association between queries and GPU IDs, the query and list IDs are forwarded to the FPGA coordination process~\ballnumber{4}, which broadcasts them to the FPGA-based disaggregated memory nodes~\ballnumber{5}.
The ChamVS near-memory processor on each node then uses the query vectors to construct distance lookup tables for each IVF list, computes the distances between the query and quantized database vectors, and collects the K nearest neighbors~\ballnumber{6}. 
Subsequently, the result vector IDs and distances from all memory nodes are sent back to the CPU server~\ballnumber{7}, which aggregates the results~\ballnumber{8} and returns the tokens of the nearest neighbors to the originating GPU~\ballnumber{9}.
Finally, the GPU predicts the next token based on both the context and the retrieved tokens~\ballnumber{10}.

\section{ChamVS Near-Memory Accelerator}

\begin{figure}[t]
  \centering
  \includegraphics[width=1.0\linewidth]{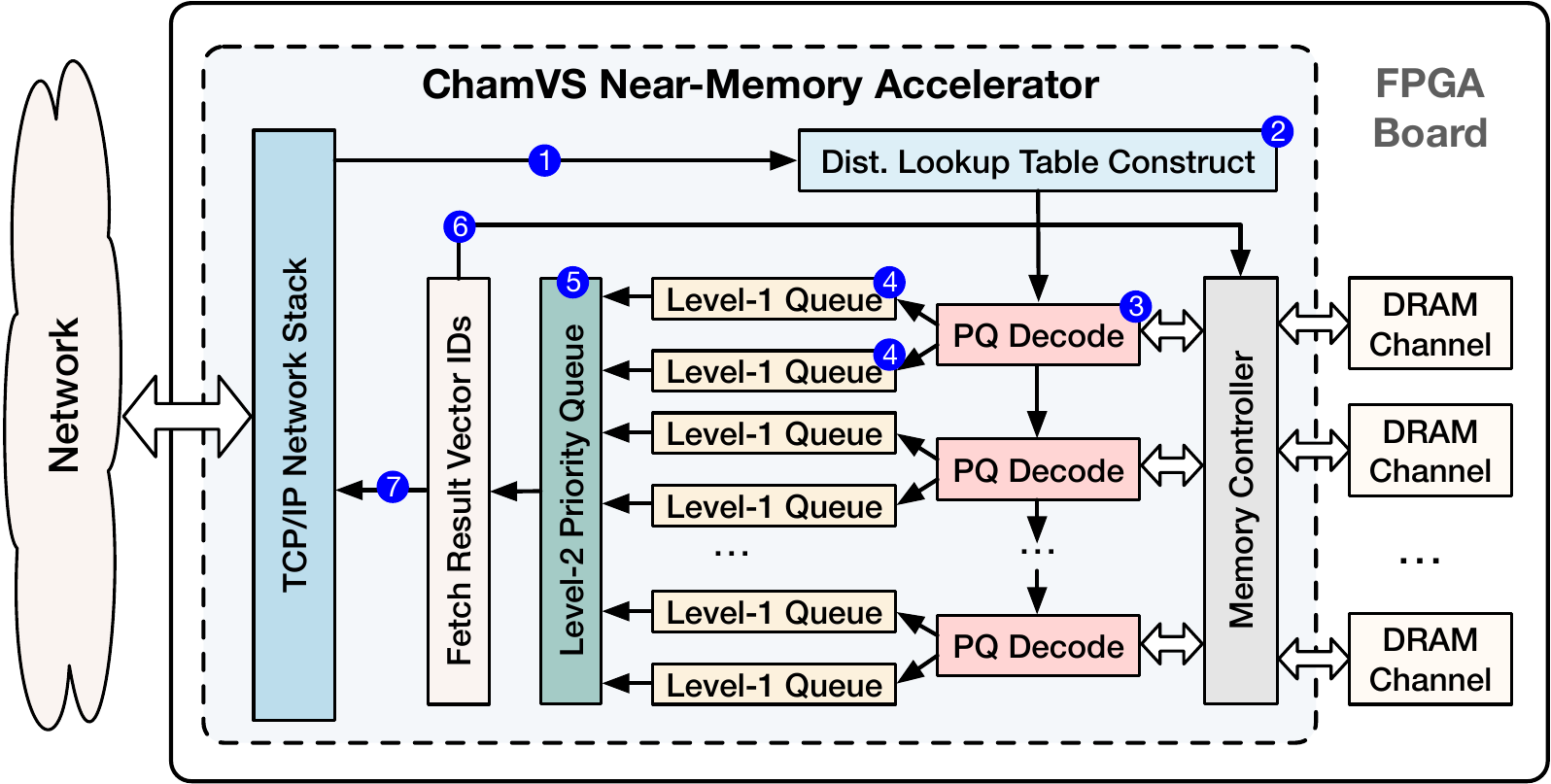}
  \vspace{-1.5em}
  \caption{The ChamVS near-memory retrieval accelerator.}
  \vspace{-1.5em}
  \label{fig:fpga-design}
\end{figure}

ChamVS enables high-performance, large-scale vector search by pairing each disaggregated memory node with a near-memory retrieval accelerator. 
As shown in Figure~\ref{fig:fpga-design}, the accelerator comprises a distance lookup table construction unit, several PQ decoding units for distance evaluations between query vectors and quantized database vectors, a group of systolic priority queues for parallel $K$-selection, and multiple memory channels.

\subsection{PQ Decoding Units}

As shown in Figure~\ref{fig:fpga-design}~\blueballnumber{3}, each ChamVS accelerator contains multiple PQ decoding units to fully utilize the memory bandwidth. These units read database vectors (PQ codes) from DRAM and compute their distances to query vectors using a distance lookup table.

\begin{figure}[t]
  \centering
  \includegraphics[width=0.8\linewidth]{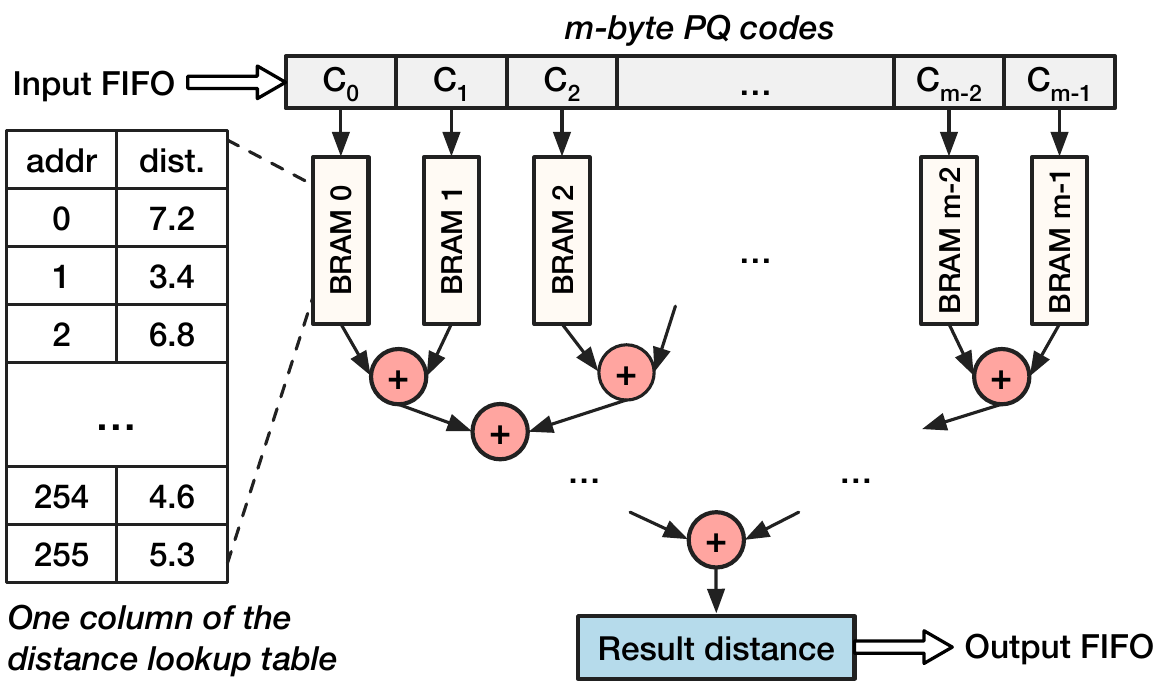}
  \vspace{-1em}
  \caption{The architecture design of a PQ decoding unit.}
  \vspace{-1.5em}
  \label{fig:PE_scan}
\end{figure}

\textbf{The design of a PQ decoding unit involves both operator and pipeline parallelisms, enabling a high throughput of producing one result distance every clock cycle.}
As shown in Figure~\ref{fig:PE_scan}, the decoding steps --- including data ingestion, distance lookups, computation, and output egestion --- are fully pipelined, similar to that of~\cite{jiang2023co, lee2022anna}. 
The unit also parallelizes the operators within the distance lookup and computation steps.

\textit{Decoding procedure.} For each IVF list to scan, the unit first stores the input distance lookup table in BRAM (on-chip SRAM in FPGAs). 
The shape of the lookup table is $m\times256$ for the typical 8-bit PQ codes ($2^8=256$), where $m$ is the number of bytes per quantized vector. Different table columns are stored in separate BRAM slices, facilitating parallel distance lookups. 
Subsequently, the PQ codes are loaded from DRAM to the unit via an $m$-byte-wide FIFO, with each byte serving as an address to retrieve a value from the corresponding column of the table. 
Finally, an adder tree sums up the retrieved values to produce the approximate distance between the query vector and the quantized database vector.

\subsection{Efficient $K$-Selection Module}

\red{The $K$-Selection module in ChamVS selects the $K$ nearest neighbors from distances computed by the PQ decoding units.}
Designing an efficient $K$-selection microarchitecture is challenging, because it has to handle multiple incoming elements per cycle due to the high throughput of PQ decoding units. 
We propose approximate hierarchical priority queue (AHPQ), a high-throughput, resource-efficient architecture for parallel $K$-selection in hardware.

\subsubsection{Primitive: Systolic Priority Queue}


The systolic priority queue facilitates high-throughput input ingestion on hardware accelerators~\cite{huang2014scalable, leiserson1979systolic}, consuming \textit{one input element every two clock cycles}.
In short, it is a register array equipped with compare-swap units between the registers, thus \textit{the hardware resource consumption of the queue increases linearly with its length.} 



A natural approach to implement $K$-selection in ChamVS is to instantiate a group of systolic priority queues in a hierarchical structure, as shown in Figure~\ref{fig:fpga-design}~\blueballnumber{4}\blueballnumber{5}.
Since a systolic priority queue can only ingest one input every two cycles, two queues, termed as level-one (L1) queues, should be paired with one PQ decoding unit, as it can produce one output per cycle. 
For each query, each L1 queue collects a subset of the $K$ nearest neighbors, and the level-two (L2) queue subsequently selects the final $K$ results.


Unfortunately, a straightforward implementation of the hierarchical priority queue can consume excessive hardware resources, making the solution unaffordable even on high-end FPGAs. 
For example, given 32 instantiated PQ decoding units and $K=100$, the accelerator would necessitate 64 L1 queues of length 100, an amount that already exceeds the total the total available FPGA resources.

\subsubsection{Approximate Hierarchical Priority Queue (AHPQ)} 
\textbf{We propose the AHPQ architecture for high-performance and resource-efficient $K$-selection.} Recognizing that ANN search is inherently approximate, we relax the $K$-selection objective from selecting the $K$ smallest distances in all queries to collecting precise results in the vast majority of cases, such as in 99\% of the queries.

\begin{figure}[t]
  \centering
  \includegraphics[width=0.95\linewidth]{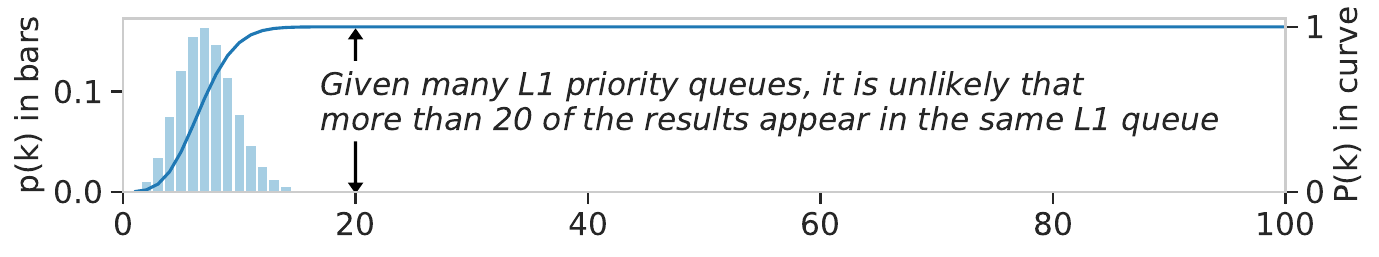}
  \vspace{-1.5em}
  \caption{The probability distribution that one out of the 16 L1 priority queues holds k out of the 100 nearest neighbors.}
  \vspace{-1.5em}
  \label{fig:single-queue-prob}
\end{figure}

The intuition behind AHPQ is simple: \textit{it is unlikely that all $K$ results are produced by a single PQ decoding unit.}
For example, given 16 level-one queues of length $K$=100, the average number of the results in a queue is only $100/16=6.25$.
Specifically, the probability that one queue holds $k$ of the $K$ results is $p(k) = C_{K}^{k} * (\frac{1}{num_{queue}})^{k} * (1-\frac{1}{num_{queue}})^{K-k}$, where $C_{K}^{k}$ represents the number of combinations selecting $k$ out of $K$ items.
The cumulative probability that a queue contains no more than $k$ of the $K$ results is $P(k) = \sum_{i=0}^{k}p(i)$.
Figure~\ref{fig:single-queue-prob} shows the probability distribution of $p$ and $P$ given different $k$ in bars and curve: it is almost impossible that a queue holds more than 20 out of the \textit{K=100} results. Thus, the lengths of the L1 queues can be truncated to 20 while producing almost the same results.  



Our design aims to reduce the size of the L1 queues while ensuring that the results for 99\% of queries remain identical to those obtained with an exact $K$-selection module. 
Specifically, for 99\% of the queries, none of the L1 queues will omit any result that is supposed to be returned to the user.

\begin{figure}[t]
  \centering
  \includegraphics[width=0.95\linewidth]{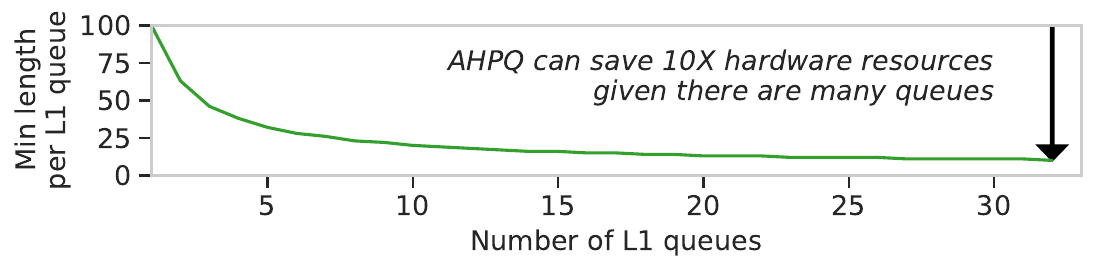}
  \vspace{-1.5em}
  \caption{The proposed approximate hierarchical priority queue can save hardware resources by an order of magnitude.}
  \vspace{-1.5em}
  \label{fig:queue-len}
\end{figure}

Figure~\ref{fig:queue-len} shows the resource savings achieved by applying the approximate hierarchical priority queue. 
As the number of L1 queues increases, the queue sizes can be reduced by an order of magnitude while still retaining 99\% of identical results, leading to a corresponding decrease in hardware resource consumption.

\subsection{Memory Management and Load Balancing}
\label{sec:memory_management}


\red{The memory management mechanism of ChamVS balances workloads across memory nodes and channels.
In our current implementation, vectors within each IVF list are evenly partitioned among memory nodes, with these sub-lists further distributed across memory channels to ensure workload balance.
For potential scenarios where IVF lists are too small to be partitioned, each list may reside on different nodes or channels, which could lead to load imbalances, especially with small query batches. Such imbalances can be mitigated by larger batches, as it is less likely that all queries happen to hit the same node or channel. 
Additionally, for the case of uneven access frequencies across IVF lists, adjusting their placement based on these frequencies can help achieve better load balancing~\cite{chen2021spann}.
}

\section{Implementation}
\label{sec:implementation}

\red{Chameleon is implemented in 11K lines of code, including 3K lines of Vitis HLS C/C++ for the ChamVS near-memory accelerator, 1.4K lines of C++ for the CPU coordinator, 3.5K lines of Python for ChamLM, and 3.2K lines of Python for various evaluations. 
}
\red{ 
Referring to existing RALM research~\cite{khandelwal2019generalization, khandelwal2020nearest}, we build ChamLM on Fairseq~\cite{ott2019fairseq}, a PyTorch-based LLM toolkit.
ChamLM extends Fairseq to support multi-GPU inference, initiating retrieval requests, integrating the retrieved tokens into generation processes, and network communication between the retrieval engines and GPU processes.
}
\red{ 
ChamVS.idx uses Faiss~\cite{johnson2019billion} for index scanning on GPUs or CPUs. 
ChamVS.mem integrates an FPGA TCP/IP stack~\cite{100gbps}. 
}
\red{ 
The CPU coordinator process for query broadcasting and result aggregation is implemented in C++ using the socket library. The simple messages in RALMs allow us to avoid higher-level abstractions like RPCs, minimizing performance overhead.
}

\section{Evaluation}
\label{sec:eval}

We evaluate Chameleon to answer the following questions:

\begin{itemize}[leftmargin=*]
    \item How much performance and energy benefits can ChamVS attain in large-scale vector search? \S~\ref{sec:eval_chamvs}
    \item How does Chameleon perform across different RALMs by introducing heterogeneous accelerators? \S~\ref{sec:eval_e2e}
    \item Is accelerator disaggregation necessary? \S~\ref{sec:eval_e2e}
\end{itemize}



\subsection{Experimental Setup}



\textbf{LLMs.} We evaluate RALM models of similar sizes to those in existing RALM research~\cite{borgeaud2022improving, li2022decoupled, sachan2021end, yogatama2021adaptive, izacard2020leveraging}, up to several billions of parameters.
We evaluate both smaller (S) and larger (L) decoder-only (Dec) and encoder-decoder (EncDec) models. Table~\ref{tab:models} summarizes the four RALMs for evaluation, including input dimensionalities, numbers of layers and attention heads, model sizes, retrieval intervals, and neighbor numbers. For EncDec models, we follow~\cite{borgeaud2022improving} to use a two-layer shallow encoder and a deeper decoder, and set different retrieval intervals. 
For all the models, we use a vocabulary size of 50K and let them generate 512 tokens per sequence. 

\textbf{Vector datasets.} 
Table~\ref{tab:datasets} summarizes the four evaluated vector datasets. 
The SIFT and Deep datasets are popular benchmarks for billion-scale ANN. 
Due to the lack of available datasets for RALM, we create two synthetic datasets by replicating each SIFT vector to the models' dimensionalities (512 and 1024). 
As a common practice, we set $nlist$, the number of clusters in the IVF index, to approximately the square root of the number of dataset vectors (\textit{nlist=32K}). We set $nprobe$ as 32 to scan 0.1\% of database vectors per query, for which high recall can be achieved on both real-world datasets (93\% on Deep and 94\% on SIFT for 100 nearest neighbors). We quantize the SIFT and Deep datasets to 16-byte PQ codes, while the two synthetic datasets adopt 32 and 64-byte PQ codes, respectively.

\textbf{Software.}
For vector search, we use \textit{Faiss}~\cite{faiss} developed by Meta, known for its optimized PQ implementations for both CPUs and GPUs. Due to its vector-only nature, Faiss's ANN search performance surpasses vector data management systems that support additional relational data functionalities~\cite{pan2023survey}. 
For LLM inference, we extend Fairseq~\cite{ott2019fairseq} to support RALMs as introduced in \S\ref{sec:implementation}.


\textbf{Hardware.} 
We instantiate the ChamVS near-memory accelerator on AMD Alveo U250 FPGAs (16 nm) equipped with 64 GB of DDR4 memory (4 channels x 16 GB) and set the accelerator frequency to 140 MHz.
For a fair comparison, each ChamVS memory node is compared to a CPU-based vector search system with equivalent memory capacity (64 GB) and an 8-core AMD EPYC 7313 processor (7 nm) with a base frequency of 3.0 GHz. 
We evaluate NVIDIA RTX 3090 GPUs (8nm) with 24 GB GDDR6X memory. 

\begin{table}
  \begin{center}
    \caption{Various RALM configurations in the evaluation.}
    \vspace{-1em}
    \label{tab:models}
    \scalebox{0.85} {
    \begin{tabular}{L{4.5em} R{2.5em} R{2.5em} R{2.5em} R{2.5em} R{3em} R{2em}} 
      \toprule
       & \multicolumn{1}{c}{Dim.}  & \multicolumn{1}{c}{Layers} & 
 \multicolumn{1}{c}{Heads} &  \multicolumn{1}{c}{Param.}  &  \multicolumn{1}{c}{Interval}  &  \multicolumn{1}{c}{$K$} \\
      \midrule
    Dec-S & 512 & 24 & 8 & 101M & 1 & 100 \\
    Dec-L &  1024 & 96 & 16 & 1259M & 1 & 100 \\
    EncDec-S &  512 & 2,24 & 8 & 158M & 8/64/512 & 10 \\
    EncDec-L &  1024 &  2,96 & 16 & 1738M & 8/64/512 & 10\\
      \bottomrule
    \end{tabular}
    } 
  \end{center}
	\vspace{-1.5em} 
\end{table}

\begin{table}
  \begin{center}
    \caption{The vector datasets used in the evaluation.}
    \vspace{-1em}
    \label{tab:datasets}
    \scalebox{0.78} {
    \begin{tabular}{L{9em} R{4em} R{4em} R{4em} R{4em}} 
      \toprule
       & \multicolumn{1}{c}{Deep} & \multicolumn{1}{c}{SIFT} &  \multicolumn{1}{c}{SYN-512} & \multicolumn{1}{c}{SYN-1024} \\
      \midrule
      \#vec & 1E+9 & 1E+9 & 1E+9 & 1E+9 \\
      $m/D$ & 16/96 & 16/128  & 32/512 & 64/1,024  \\
      \textit{nprobe/nlist} & 32/32K & 32/32K & 32/32K & 32/32K \\
      Raw vectors (GB) & 384 & 512 & 2,048 & 4,096 \\
      PQ and vec IDs (GB) & 24 & 24 & 40 & 72 \\
      \bottomrule
    \end{tabular}
    } 
  \end{center}
	\vspace{-1em} 
\end{table}

\begin{figure*}[t]
  
  \centering

  \begin{subfigure}[b]{0.24\linewidth}
    \includegraphics[width=\linewidth]{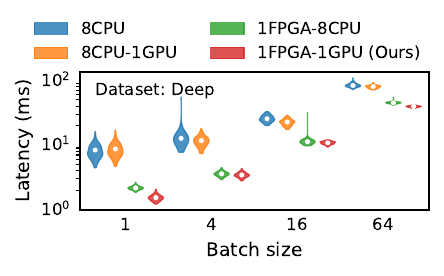}
  \end{subfigure}
  \begin{subfigure}[b]{0.24\linewidth}
    \includegraphics[width=\linewidth]{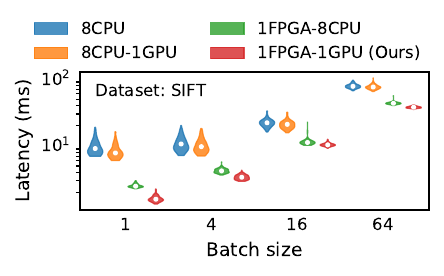}
  \end{subfigure}
  \begin{subfigure}[b]{0.24\linewidth}
    \includegraphics[width=\linewidth]{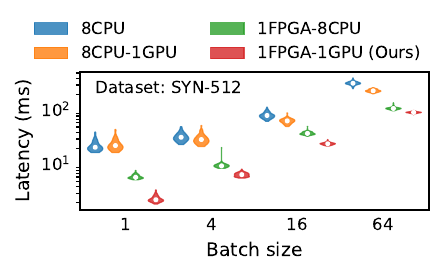}
  \end{subfigure}
  \begin{subfigure}[b]{0.24\linewidth}
    \includegraphics[width=\linewidth]{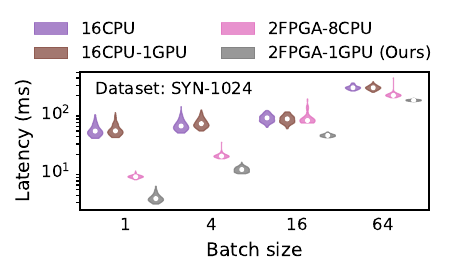}
  \end{subfigure}
  \hfill
  
  \vspace{-1.5em}
  \caption{ChamVS achieves significantly lower vector search latency than CPUs and GPUs.}
  \vspace{-1.5em}
  \label{fig:chamvs_latency}
\end{figure*}

  

  
  


\subsection{Large-Scale Vector Search on ChamVS}
\label{sec:eval_chamvs}

\textbf{Search performance.} 
We compare ChamVS with baseline systems using four hardware setups. PQ codes can be processed on CPU/FPGA while the IVF index can be scanned on CPU/GPU, leading to four hardware configurations: CPU, CPU-GPU, FPGA-CPU, and FPGA-GPU.
To report the best baseline performance, the CPU and CPU-GPU systems are monolithic, while the FPGA-CPU and FPGA-GPU systems are disaggregated over the network. 
Figure~\ref{fig:chamvs_latency} compares the latency distributions of the four solutions. Each white dot in the violin plots denotes a median latency. The number of CPU cores and the number of accelerators used are listed in the plot legends. We make two primary observations from the experiments:


\textit{Firstly, the near-memory accelerator in ChamVS significantly lowers vector search latency.} Across different datasets and batch sizes (Figure~\ref{fig:chamvs_latency}), the FPGA-CPU solution achieves 1.36$\sim$6.13$\times$ speedup compared to the CPU baseline, and the FPGA-GPU solution shows even higher speedup (2.25$\sim$23.72$\times$). This is because the ChamVS near memory accelerator can (a) decode PQ codes in parallel, (b) pipeline the decoding, distance calculation, and K-selection, such that each quantized vector can be processed by the pipeline rapidly.

\textit{Secondly, scanning the IVF index on GPU allows further latency improvements compared to the FPGA-CPU solution.} 
As shown in Figure~\ref{fig:chamvs_latency}, the FPGA-GPU approach achieves 1.04$\sim$3.87$\times$ speedup compared to the FPGA-CPU solution. This is because the IVF index scan procedure can easily leverage the massively parallelism and the high memory bandwidth of GPUs. 
In contrast, the hybrid CPU-GPU solution shows little or even negative improvements compared to the CPU-only solution (0.91$\sim$1.42$\times$), because the search performance is limited by the slow PQ code scan process on CPU.

\begin{figure}[t]
  \centering
  \includegraphics[width=0.85\linewidth]{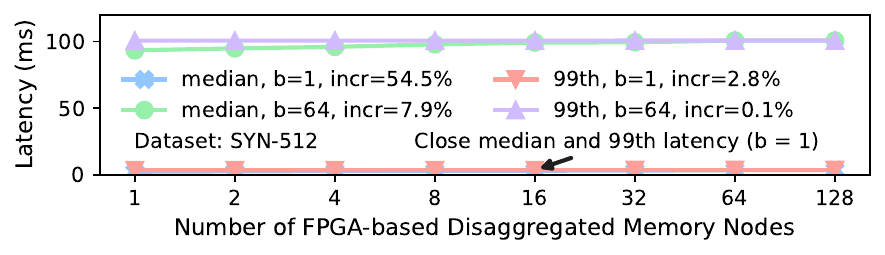}
  \vspace{-1.5em}
  \caption{\red{The performance scalability of ChamVS.}}
  \vspace{-1.5em}
  \label{fig:scalability}
\end{figure}

\textbf{Scalability.}
We extrapolate query latency beyond the limited number of accelerators available in our evaluation. Considering the one-GPU and $N$-FPGA setup, we estimate the latency distribution by summing up accelerator and network latencies. Each query latency number is the maximum of $N$ randomly sampled latency numbers from the 1-FPGA setup. For network latency, we assume a 100 Gbps bandwidth for the CPU server and apply the LogGP model~\cite{alexandrov1995loggp, culler1993logp}, which assumes a tree topology for broadcast and reduce communications, setting the latency between two endpoints as 10.0 $\mu$s (a conservative number compared to 6.0 $\mu$s reported in~\cite{hoefler2007low, hoefler2014energy}). 
Figure~\ref{fig:scalability} presents the median and the 99th percentile latencies for different batch sizes on the SYN-512 dataset. The tail latencies remain almost identical to those in the one-node setup due to the negligible network latency compared to the query. As for the median latencies, there is only a 7.9\% increase for a batch size of 64, while for the case without batching, the latency increases by 54.5\% as the accelerator latency is determined by the slowest one.

\textbf{Energy consumption.} \textit{ChamVS achieves 5.8$\sim$26.2$\times$ energy efficiency compared to the CPU.} Table~\ref{tab:energy} summarizes the average energy consumption to serve a single query across different systems. We measure CPU, GPU, and FPGA energy consumption using Running Average Power Limit (RAPL) and NVIDIA System Management Interface, and Vivado, respectively. For ChamVS, we report the energy per query by measuring the power consumption times latency for scanning index on GPU and scanning PQ codes on FPGAs, respectively, and summing the two parts up.

\begin{table}
\begin{small}
  \begin{center}
    \caption{Average energy consumption per query (in mJ) on ChamVS and CPUs using various batch sizes (1$\sim$16).}
    \vspace{-2em}
    \label{tab:energy}
    \begin{small}
    \scalebox{0.8} {
    \begin{tabular}{L{5em}     M{3em} M{0em} M{2em} M{0em} M{2em} M{0em}      M{2em} M{0em} M{2em} M{0em} M{2em}   }\\
      \toprule
      & \multicolumn{5}{c}{CPU} & \phantom{}& \multicolumn{5}{c}{ChamVS (FPGA + GPU)} \\
        \cmidrule{2-6} \cmidrule{8-12}
      & b=1 & \phantom{}& b=4 & \phantom{}& b=16 & \phantom{}&   b=1 & \phantom{}& b=4 & \phantom{}& b=16  \\
      \midrule
SIFT &  950.3 && 434.0 && 143.3 && 53.6 && 28.2 && 21.5 \\
Deep &  929.5 && 412.9 && 141.9 && 52.3 && 26.9 && 20.5 \\
SYN-512 &  1734.9 && 957.8 && 372.5 && 95.6 && 55.0 && 41.1 \\
SYN-1024 &  4459.9 && 2315.0 && 918.5 && 170.1 && 107.8 && 85.2 \\
      \bottomrule
    \end{tabular}
    } 
    \end{small}
  \end{center}
\end{small}
  \vspace{-1em}
\end{table}



\red{\textbf{Recall}. 
\textit{ChamVS, with approximate hierarchical priority queues (AHPQ), delivers results nearly identical to those of the software.}
Table~\ref{tab:recall} shows the recall given various AHPQ lengths (8$\sim$32) when searching for the $K=100$ nearest neighbors. Here, R1@100 indicates the percentage of queries where the top nearest neighbor is within the results, while R@100 represents the percentage of overlap between the true 100 nearest neighbors and the 100 results returned.
Compared to software, AHPQ only decreases recall by up to 0.06\%. 
Interestingly, on the Deep dataset, reducing the queue lengths to eight does not necessarily result in lower recall than using a length of 32. This is likely due to the nature of PQ approximation --- a higher distance indicated by PQ does not always mean that the original vector is actually farther from the query.
}

\begin{table}
\begin{small}
  \begin{center}
    \caption{\red{Recall of ChamVS using approximate queues.}}
    \vspace{-2em}
    \label{tab:recall}
    \begin{small}
    \scalebox{0.8} {
    \begin{tabular}{L{6.5em}    R{6em} R{6.2em}   R{6.2em}  R{6.2em} }\\
      \toprule
      & CPU (len=100) & AHPQ (len=8) &   AHPQ (len=16) & AHPQ (len=32)  \\
      \midrule

R1@100 (Deep) &  92.88\% & 92.85\%  &  92.84\% & 92.84\% \\
R@100 (Deep) &  45.54\% & 45.49\% &  45.49\% &  45.48\% \\
R1@100 (SIFT) &  94.21\% & 94.20\% &  94.21\% & 94.21\% \\
R@100 (SIFT) &  48.68\% & 48.66\% &  48.67\% & 48.67\% \\

      \bottomrule
    \end{tabular}
    } 
    \end{small}
  \end{center}
\end{small}
  \vspace{-1em}
\end{table}

\subsection{End-to-end RALM Inference on Chameleon}
\label{sec:eval_e2e}

We evaluate RALM inference performance on Chameleon with different models and retrieval intervals, using the SYN-512 and SYN-1024 datasets for the smaller and larger models, respectively. 

\textbf{RALM performance.} 
We evaluate system performance when generating a 512-token sequence using a single GPU for LLM inference .
For the latency evaluation, we disable batching, while the throughput evaluation uses the maximum allowed batch sizes given the GPU's memory capacity (64 for Dec-S and EncDec-S; 8 for Dec-L and EncDec-L).
For vector search in RALM, we use the FPGA-GPU solution for ChamVS and the CPU-only solution as the baseline, as CPU-GPU vector search can be even slower using small batches.

\begin{figure}[t]  
  \centering
    \includegraphics[width=1.0\linewidth]{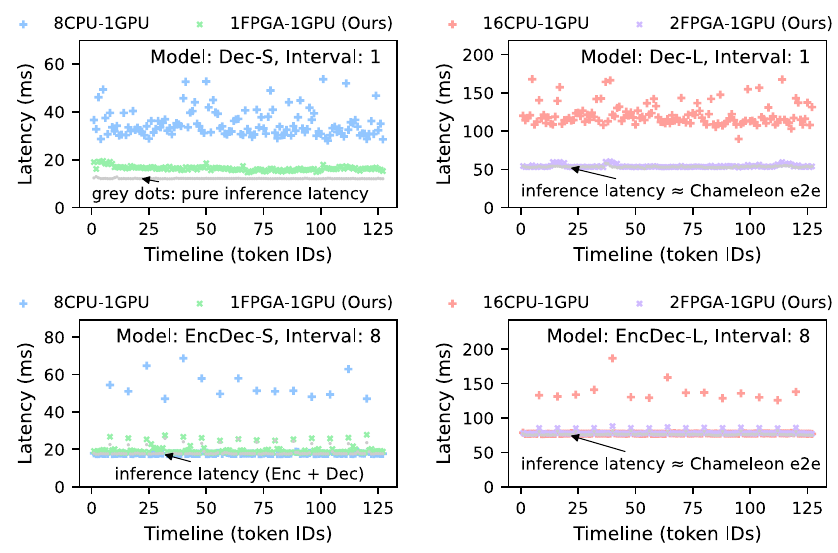}
  \vspace{-1.5em}
  \caption{\red{Latency of RALM inference given different LLM configurations and retrieval intervals.}}
  \vspace{-1.5em}
  \label{fig:chameleon_latency_over_time}
\end{figure}

\textit{Chameleon significantly outperforms the CPU-GPU baseline system in latency for inference steps involving vector search.}
\red{Figure~\ref{fig:chameleon_latency_over_time} visualizes the RALM inference latency of Chameleon and the baseline system (CPU-GPU) for the first 128 generated tokens. 
Inference latency is represented by the grey dots, while retrieval latency accounts for the remaining portion of the end-to-end latency. The time spent on coordinator and index scanning is not marked in the figure, as their latencies of hundreds of microseconds are negligible compared to up to tens of milliseconds for inference and retrieval. }
\red{Figure~\ref{fig:chameleon_latency_over_time} shows that ChamVS significantly reduces the latency at the token generation steps requiring retrieval, as the retrieval latency of Chameleon is almost negligible compared to the inference latency executed on GPUs. }
Specifically, the speedup provided by Chameleon at retrieval-based inference steps (retrieval + inference) ranges from 1.94$\sim$4.11$\times$, 1.71$\sim$3.02$\times$, 1.76$\sim$3.41$\times$, and 1.29$\sim$2.13$\times$ for Dec-S, EncDec-S, Dec-L, and EncDec-L, respectively.

\begin{figure}[t]
  \centering

  \begin{subfigure}[b]{0.49\linewidth}
    \includegraphics[width=\linewidth]{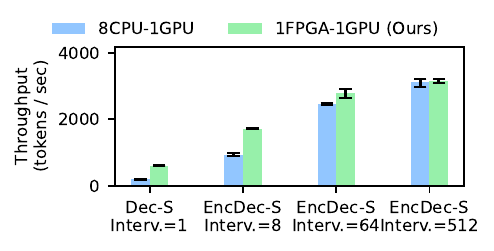}
  \end{subfigure}
  \begin{subfigure}[b]{0.49\linewidth}
    \includegraphics[width=\linewidth]{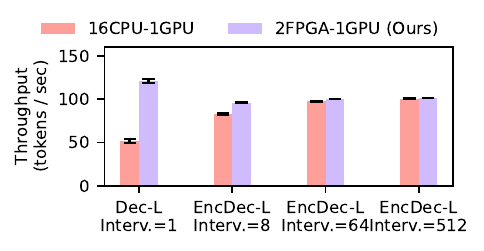}
  \end{subfigure}
  \hfill
  
  \vspace{-1.5em}
  \caption{Throughput of RALM inference given different LLM configurations and retrieval intervals.}
  \label{fig:chameleon_throughput}
  \vspace{-1.5em}
\end{figure}

\textit{Chameleon achieves up to 3.18$\times$ throughput compared to the CPU-GPU baseline.} Figure~\ref{fig:chameleon_throughput} shows that the lower the retrieval interval, the more throughput advantage Chameleon offers, with the speedup being 3.18$\times$ and 2.34$\times$ for Dec-S and Dec-L that require retrieval per token generation (interval=1). 
Chameleon attains greater speedup in batched inference than single-sequence inference (as in latency experiments), because, as the batch size grows, the latency increase for LLM inference is not as significant as that of vector search, due to the many-core parallelism that GPUs offer. 

\begin{figure}[t]
  \centering
  \includegraphics[width=0.85\linewidth]{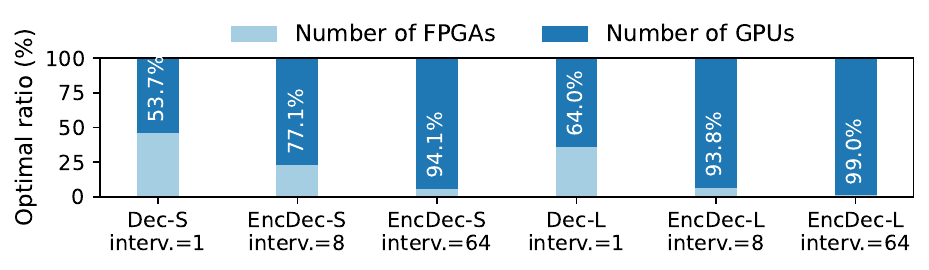}
  \vspace{-1.5em}
  \caption{\red{Disaggregation is essential as the optimal accelerator ratio varies significantly across RALM configurations.}}
  \vspace{-1.5em}
  \label{fig:accelerator_ratio}
\end{figure}

\red{\textbf{The need for resource disaggregation.} 
Accelerator disaggregation allows Chameleon to adjust the ratio between the two types of accelerators across RALM configurations. 
We model the overall system throughput, measured by generated tokens per second, across various accelerator ratios using a total of 1,000 accelerators, assuming the cost for an inference accelerator and a retrieval accelerator is equivalent.  Given retrieval interval \(i\), batch size \(b\), number of inference and retrieval accelerators \(N_I\) and \(N_R\), latency per batch for inference and retrieval \(L_I(b)\) and \(L_R(b)\), the system throughput is determined by the minimum of the inference and retrieval throughput: \( Th_{system} = \min(Th_{I}, Th_{R}) \), where \( Th_{I} = \frac{i \cdot b \cdot N_I}{i \cdot L_I(b) + L_R(b)}\) and \( Th_{R} = \frac{i \cdot b \cdot N_R }{L_R(b)} \).
}
\red{
Figure~\ref{fig:accelerator_ratio} shows that the optimal ratio of accelerators to achieve the highest throughput varies significantly, ranging from 53.7\%$\sim$99.0\% across RALMs.}

\begin{figure}[t]
  \centering
  \includegraphics[width=1.0\linewidth]{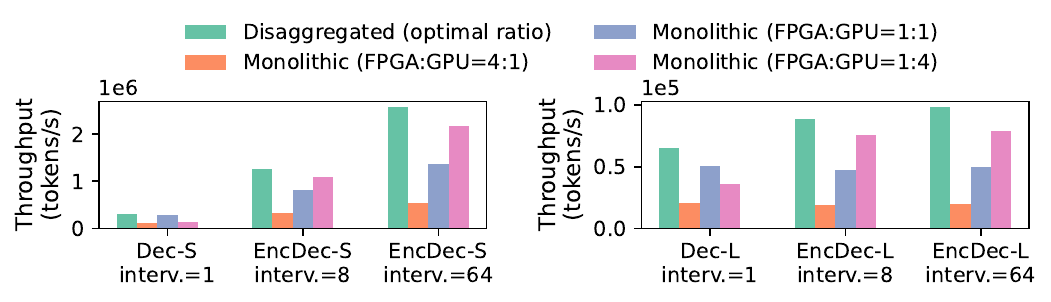}
  \vspace{-1.5em}
  \caption{\red{The disaggregated design consistently outperforms the monolithic ones using fixed accelerator ratios.}}
  \vspace{-1.5em}
  \label{fig:disaggregated_vs_monolithic}
\end{figure}

\red{
\textit{The disaggregated design, using the optimal accelerator ratio, consistently outperforms the monolithic ones with fixed ratios, as shown in Figure~\ref{fig:disaggregated_vs_monolithic}.}
Given the impracticality of adjusting the ratio for each RALM in a monolithic design, the performance of a monolithic design can only match that of Chameleon on a limited set of RALMs.
}

\section{Related Work}
\label{sec:related_work}

To our knowledge, Chameleon represents the first endeavor to improve RALM inference performance by using heterogeneous accelerator systems. We now introduce related research topics below.

\textbf{ANN search.} Researchers have developed various ANN search algorithms~\cite{huang2021point, ouyang2020progressive, zhu2016range, zheng2016lazylsh, wang2015optimal, jiang2015exact, sun2014srs, dallachiesa2014top, yang2015reverse, lu2012efficient, gao2024rabitq} and data management systems~\cite{wang2021milvus, guo2022manu, adb-v, mohoney2023high, yang2020pase, pan2023survey}.
Apart from PQ-based vector search, graph-based searching algorithms~\cite{malkov2014approximate, malkov2018efficient, fu2017fast, wu2014fast, zhao2023towards, zuo2023arkgraph, lu2021hvs, peng2023efficient, gao2023high} are popular as they can achieve high recall and low latency.
Locality-sensitive hashing (LSH)~\cite{gionis1999similarity, datar2004locality} offers theoretical guarantees in ANN search but empirically does not perform as well as PQ and graph-based algorithms. 

\textbf{Vector search on modern hardware.}
Faiss is the most popular GPU-accelerated ANN search library so far~\cite{johnson2019billion}, and there are several academic GPU implementations~\cite{wieschollek2016efficient, chen2019robustiq, chen2019vector, liu2023juno}. 
Lee et al.~\cite{lee2022anna} study ASIC designs for IVF-PQ, and a couple of works~\cite{jiang2023co, zhang2018efficient} implement IVF-PQ on an FPGA, but their designs are constrained by either the limited HBM capacity or the slow CPU-FPGA interconnect. 
In contrast, Chameleon disaggregates IVF-PQ, with the index on GPUs and PQ codes on FPGA-based memory nodes, and employs the innovative hardware priority queue design to achieve high performance with little hardware resources. 
While graph-based vector search accelerators can achieve low latency~\cite{jiang2024accelerating, zeng2023df}, the memory consumption is high, requiring up to one TB of memory for only one billion SIFT vectors, in contrast to 24 GB in our case. 
Apart from accelerators, researchers also study memory and storage for vector search. 
One can leverage non-volatile memory~\cite{ren2020hm} and CXL~\cite{jang2023cxl} to scale up graph-based ANN, while on-disk ANN has to be more careful with I/O cost~\cite{chen2021spann, jayaram2019diskann, lejsek2008nv}. 
Hu et al.~\cite{hu2022ice} further push down distance evaluation into NAND flash to reduce data movement.

\section{Conclusion and Outlook}
\label{sec:discussion}


We present Chameleon, a heterogeneous and disaggregated accelerator system for efficient RALM inference. 
Given the rapidly evolving algorithms, software, and hardware related to RALMs, Chameleon can be potentially upgraded in the following ways.
For LLM inference, ChamLM could be enhanced by supporting low precision~\cite{fastertransformer}, continuous batching~\cite{yu2022orca}, paged-attention~\cite{kwon2023efficient}, and disaggregated prompt computation and token generation~\cite{patel2023splitwise}. 
While currently supporting PQ, ChamVS could potentially be replaced by graph-based ANN accelerators~\cite{zeng2023df, peng2021optimizing}. 
ChamVS could also be extended to support index updates~\cite{xu2023spfresh} and relational features~\cite{zhang2023vbase}.






\begin{acks} 
We thank AMD for their generous donation of the Heterogeneous Accelerated Compute Clusters (HACC) at ETH Zurich (\url{https://systems.ethz.ch/research/data-processing-on-modern-hardware/hacc.html}), on which the experiments were conducted. 
\end{acks}


\bibliographystyle{ACM-Reference-Format}
\bibliography{ref}


\begin{thebibliography}{94}


\ifx \showCODEN    \undefined \def \showCODEN     #1{\unskip}     \fi
\ifx \showDOI      \undefined \def \showDOI       #1{#1}\fi
\ifx \showISBNx    \undefined \def \showISBNx     #1{\unskip}     \fi
\ifx \showISBNxiii \undefined \def \showISBNxiii  #1{\unskip}     \fi
\ifx \showISSN     \undefined \def \showISSN      #1{\unskip}     \fi
\ifx \showLCCN     \undefined \def \showLCCN      #1{\unskip}     \fi
\ifx \shownote     \undefined \def \shownote      #1{#1}          \fi
\ifx \showarticletitle \undefined \def \showarticletitle #1{#1}   \fi
\ifx \showURL      \undefined \def \showURL       {\relax}        \fi
\providecommand\bibfield[2]{#2}
\providecommand\bibinfo[2]{#2}
\providecommand\natexlab[1]{#1}
\providecommand\showeprint[2][]{arXiv:#2}

\bibitem[\protect\citeauthoryear{??}{fai}{[n.d.]}]%
        {faiss}
 \bibinfo{year}{[n.d.]}\natexlab{}.
\newblock \bibinfo{title}{Faiss}.
\newblock \bibinfo{howpublished}{\url{https://github.com/facebookresearch/faiss /}}.
\newblock


\bibitem[\protect\citeauthoryear{??}{fas}{[n.d.]}]%
        {fastertransformer}
 \bibinfo{year}{[n.d.]}\natexlab{}.
\newblock \bibinfo{title}{FasterTransformer}.
\newblock \bibinfo{howpublished}{\url{ https://github.com/NVIDIA/FasterTransformer}}.
\newblock


\bibitem[\protect\citeauthoryear{??}{Ope}{[n.d.]a}]%
        {OpenAI_assistant_2}
 \bibinfo{year}{[n.d.]}\natexlab{a}.
\newblock \bibinfo{title}{The Implications of OpenAI’s Latest Update on RAG and Vector-Only Databases}.
\newblock \bibinfo{howpublished}{https://medium.com/@vishalkalia.er/the-implications-of-openais-latest-update-on-rag-and-vector-only-databases-c3f326cce0a1}.
\newblock


\bibitem[\protect\citeauthoryear{??}{Ope}{[n.d.]b}]%
        {OpenAI_assistant_1}
 \bibinfo{year}{[n.d.]}\natexlab{b}.
\newblock \bibinfo{title}{What does OpenAI’s announcement mean for Retrieval Augmented Generation (RAG) and Vector-only Databases?}
\newblock \bibinfo{howpublished}{https://medium.com/madhukarkumar/what-does-openais-announcement-mean-for-retrieval-augmented-generation-rag-and-vector-only-54bfc34cba2c}.
\newblock


\bibitem[\protect\citeauthoryear{Alexandrov, Ionescu, Schauser, and Scheiman}{Alexandrov et~al\mbox{.}}{1995}]%
        {alexandrov1995loggp}
\bibfield{author}{\bibinfo{person}{Albert Alexandrov}, \bibinfo{person}{Mihai~F Ionescu}, \bibinfo{person}{Klaus~E Schauser}, {and} \bibinfo{person}{Chris Scheiman}.} \bibinfo{year}{1995}\natexlab{}.
\newblock \showarticletitle{LogGP: Incorporating long messages into the LogP model—one step closer towards a realistic model for parallel computation}. In \bibinfo{booktitle}{\emph{Proceedings of the seventh annual ACM symposium on Parallel algorithms and architectures}}. \bibinfo{pages}{95--105}.
\newblock


\bibitem[\protect\citeauthoryear{Alon, Xu, He, Sengupta, Roth, and Neubig}{Alon et~al\mbox{.}}{2022}]%
        {alon2022neuro}
\bibfield{author}{\bibinfo{person}{Uri Alon}, \bibinfo{person}{Frank Xu}, \bibinfo{person}{Junxian He}, \bibinfo{person}{Sudipta Sengupta}, \bibinfo{person}{Dan Roth}, {and} \bibinfo{person}{Graham Neubig}.} \bibinfo{year}{2022}\natexlab{}.
\newblock \showarticletitle{Neuro-symbolic language modeling with automaton-augmented retrieval}. In \bibinfo{booktitle}{\emph{International Conference on Machine Learning}}. PMLR, \bibinfo{pages}{468--485}.
\newblock


\bibitem[\protect\citeauthoryear{Borgeaud, Mensch, Hoffmann, Cai, Rutherford, Millican, Van Den~Driessche, Lespiau, Damoc, Clark, et~al\mbox{.}}{Borgeaud et~al\mbox{.}}{2022}]%
        {borgeaud2022improving}
\bibfield{author}{\bibinfo{person}{Sebastian Borgeaud}, \bibinfo{person}{Arthur Mensch}, \bibinfo{person}{Jordan Hoffmann}, \bibinfo{person}{Trevor Cai}, \bibinfo{person}{Eliza Rutherford}, \bibinfo{person}{Katie Millican}, \bibinfo{person}{George~Bm Van Den~Driessche}, \bibinfo{person}{Jean-Baptiste Lespiau}, \bibinfo{person}{Bogdan Damoc}, \bibinfo{person}{Aidan Clark}, {et~al\mbox{.}}} \bibinfo{year}{2022}\natexlab{}.
\newblock \showarticletitle{Improving language models by retrieving from trillions of tokens}. In \bibinfo{booktitle}{\emph{International conference on machine learning}}. PMLR, \bibinfo{pages}{2206--2240}.
\newblock


\bibitem[\protect\citeauthoryear{Brown, Mann, Ryder, Subbiah, Kaplan, Dhariwal, Neelakantan, Shyam, Sastry, Askell, et~al\mbox{.}}{Brown et~al\mbox{.}}{2020}]%
        {brown2020language}
\bibfield{author}{\bibinfo{person}{Tom Brown}, \bibinfo{person}{Benjamin Mann}, \bibinfo{person}{Nick Ryder}, \bibinfo{person}{Melanie Subbiah}, \bibinfo{person}{Jared~D Kaplan}, \bibinfo{person}{Prafulla Dhariwal}, \bibinfo{person}{Arvind Neelakantan}, \bibinfo{person}{Pranav Shyam}, \bibinfo{person}{Girish Sastry}, \bibinfo{person}{Amanda Askell}, {et~al\mbox{.}}} \bibinfo{year}{2020}\natexlab{}.
\newblock \showarticletitle{Language models are few-shot learners}.
\newblock \bibinfo{journal}{\emph{Advances in neural information processing systems}}  \bibinfo{volume}{33} (\bibinfo{year}{2020}), \bibinfo{pages}{1877--1901}.
\newblock


\bibitem[\protect\citeauthoryear{Chen, Zhao, Wang, Li, Liu, Li, Yang, and Wang}{Chen et~al\mbox{.}}{2021}]%
        {chen2021spann}
\bibfield{author}{\bibinfo{person}{Qi Chen}, \bibinfo{person}{Bing Zhao}, \bibinfo{person}{Haidong Wang}, \bibinfo{person}{Mingqin Li}, \bibinfo{person}{Chuanjie Liu}, \bibinfo{person}{Zengzhong Li}, \bibinfo{person}{Mao Yang}, {and} \bibinfo{person}{Jingdong Wang}.} \bibinfo{year}{2021}\natexlab{}.
\newblock \showarticletitle{SPANN: Highly-efficient Billion-scale Approximate Nearest Neighbor Search}.
\newblock \bibinfo{journal}{\emph{arXiv preprint arXiv:2111.08566}} (\bibinfo{year}{2021}).
\newblock


\bibitem[\protect\citeauthoryear{Chen, Chen, Zou, Li, Lu, Wang, and Zhao}{Chen et~al\mbox{.}}{2019b}]%
        {chen2019vector}
\bibfield{author}{\bibinfo{person}{Wei Chen}, \bibinfo{person}{Jincai Chen}, \bibinfo{person}{Fuhao Zou}, \bibinfo{person}{Yuan-Fang Li}, \bibinfo{person}{Ping Lu}, \bibinfo{person}{Qiang Wang}, {and} \bibinfo{person}{Wei Zhao}.} \bibinfo{year}{2019}\natexlab{b}.
\newblock \showarticletitle{Vector and line quantization for billion-scale similarity search on GPUs}.
\newblock \bibinfo{journal}{\emph{Future Generation Computer Systems}}  \bibinfo{volume}{99} (\bibinfo{year}{2019}), \bibinfo{pages}{295--307}.
\newblock


\bibitem[\protect\citeauthoryear{Chen, Chen, Zou, Li, Lu, and Zhao}{Chen et~al\mbox{.}}{2019a}]%
        {chen2019robustiq}
\bibfield{author}{\bibinfo{person}{Wei Chen}, \bibinfo{person}{Jincai Chen}, \bibinfo{person}{Fuhao Zou}, \bibinfo{person}{Yuan-Fang Li}, \bibinfo{person}{Ping Lu}, {and} \bibinfo{person}{Wei Zhao}.} \bibinfo{year}{2019}\natexlab{a}.
\newblock \showarticletitle{Robustiq: A robust ann search method for billion-scale similarity search on gpus}. In \bibinfo{booktitle}{\emph{Proceedings of the 2019 on International Conference on Multimedia Retrieval}}. \bibinfo{pages}{132--140}.
\newblock


\bibitem[\protect\citeauthoryear{Chowdhery, Narang, Devlin, Bosma, Mishra, Roberts, Barham, Chung, Sutton, Gehrmann, et~al\mbox{.}}{Chowdhery et~al\mbox{.}}{2022}]%
        {chowdhery2022palm}
\bibfield{author}{\bibinfo{person}{Aakanksha Chowdhery}, \bibinfo{person}{Sharan Narang}, \bibinfo{person}{Jacob Devlin}, \bibinfo{person}{Maarten Bosma}, \bibinfo{person}{Gaurav Mishra}, \bibinfo{person}{Adam Roberts}, \bibinfo{person}{Paul Barham}, \bibinfo{person}{Hyung~Won Chung}, \bibinfo{person}{Charles Sutton}, \bibinfo{person}{Sebastian Gehrmann}, {et~al\mbox{.}}} \bibinfo{year}{2022}\natexlab{}.
\newblock \showarticletitle{Palm: Scaling language modeling with pathways}.
\newblock \bibinfo{journal}{\emph{arXiv preprint arXiv:2204.02311}} (\bibinfo{year}{2022}).
\newblock


\bibitem[\protect\citeauthoryear{Culler, Karp, Patterson, Sahay, Schauser, Santos, Subramonian, and Von~Eicken}{Culler et~al\mbox{.}}{1993}]%
        {culler1993logp}
\bibfield{author}{\bibinfo{person}{David Culler}, \bibinfo{person}{Richard Karp}, \bibinfo{person}{David Patterson}, \bibinfo{person}{Abhijit Sahay}, \bibinfo{person}{Klaus~Erik Schauser}, \bibinfo{person}{Eunice Santos}, \bibinfo{person}{Ramesh Subramonian}, {and} \bibinfo{person}{Thorsten Von~Eicken}.} \bibinfo{year}{1993}\natexlab{}.
\newblock \showarticletitle{LogP: Towards a realistic model of parallel computation}. In \bibinfo{booktitle}{\emph{Proceedings of the fourth ACM SIGPLAN symposium on Principles and practice of parallel programming}}. \bibinfo{pages}{1--12}.
\newblock


\bibitem[\protect\citeauthoryear{Dallachiesa, Palpanas, and Ilyas}{Dallachiesa et~al\mbox{.}}{2014}]%
        {dallachiesa2014top}
\bibfield{author}{\bibinfo{person}{Michele Dallachiesa}, \bibinfo{person}{Themis Palpanas}, {and} \bibinfo{person}{Ihab~F Ilyas}.} \bibinfo{year}{2014}\natexlab{}.
\newblock \showarticletitle{Top-k nearest neighbor search in uncertain data series}.
\newblock \bibinfo{journal}{\emph{Proceedings of the VLDB Endowment}} \bibinfo{volume}{8}, \bibinfo{number}{1} (\bibinfo{year}{2014}), \bibinfo{pages}{13--24}.
\newblock


\bibitem[\protect\citeauthoryear{Datar, Immorlica, Indyk, and Mirrokni}{Datar et~al\mbox{.}}{2004}]%
        {datar2004locality}
\bibfield{author}{\bibinfo{person}{Mayur Datar}, \bibinfo{person}{Nicole Immorlica}, \bibinfo{person}{Piotr Indyk}, {and} \bibinfo{person}{Vahab~S Mirrokni}.} \bibinfo{year}{2004}\natexlab{}.
\newblock \showarticletitle{Locality-sensitive hashing scheme based on p-stable distributions}. In \bibinfo{booktitle}{\emph{Proceedings of the twentieth annual symposium on Computational geometry}}. \bibinfo{pages}{253--262}.
\newblock


\bibitem[\protect\citeauthoryear{Devlin, Chang, Lee, and Toutanova}{Devlin et~al\mbox{.}}{2018}]%
        {devlin2018bert}
\bibfield{author}{\bibinfo{person}{Jacob Devlin}, \bibinfo{person}{Ming-Wei Chang}, \bibinfo{person}{Kenton Lee}, {and} \bibinfo{person}{Kristina Toutanova}.} \bibinfo{year}{2018}\natexlab{}.
\newblock \showarticletitle{Bert: Pre-training of deep bidirectional transformers for language understanding}.
\newblock \bibinfo{journal}{\emph{arXiv preprint arXiv:1810.04805}} (\bibinfo{year}{2018}).
\newblock


\bibitem[\protect\citeauthoryear{Fu, Xiang, Wang, and Cai}{Fu et~al\mbox{.}}{2017}]%
        {fu2017fast}
\bibfield{author}{\bibinfo{person}{Cong Fu}, \bibinfo{person}{Chao Xiang}, \bibinfo{person}{Changxu Wang}, {and} \bibinfo{person}{Deng Cai}.} \bibinfo{year}{2017}\natexlab{}.
\newblock \showarticletitle{Fast approximate nearest neighbor search with the navigating spreading-out graph}.
\newblock \bibinfo{journal}{\emph{arXiv preprint arXiv:1707.00143}} (\bibinfo{year}{2017}).
\newblock


\bibitem[\protect\citeauthoryear{Gao and Long}{Gao and Long}{2023}]%
        {gao2023high}
\bibfield{author}{\bibinfo{person}{Jianyang Gao} {and} \bibinfo{person}{Cheng Long}.} \bibinfo{year}{2023}\natexlab{}.
\newblock \showarticletitle{High-dimensional approximate nearest neighbor search: with reliable and efficient distance comparison operations}.
\newblock \bibinfo{journal}{\emph{Proceedings of the ACM on Management of Data}} \bibinfo{volume}{1}, \bibinfo{number}{2} (\bibinfo{year}{2023}), \bibinfo{pages}{1--27}.
\newblock


\bibitem[\protect\citeauthoryear{Gao and Long}{Gao and Long}{2024}]%
        {gao2024rabitq}
\bibfield{author}{\bibinfo{person}{Jianyang Gao} {and} \bibinfo{person}{Cheng Long}.} \bibinfo{year}{2024}\natexlab{}.
\newblock \showarticletitle{RaBitQ: Quantizing High-Dimensional Vectors with a Theoretical Error Bound for Approximate Nearest Neighbor Search}.
\newblock \bibinfo{journal}{\emph{Proceedings of the ACM on Management of Data}} \bibinfo{volume}{2}, \bibinfo{number}{3} (\bibinfo{year}{2024}), \bibinfo{pages}{1--27}.
\newblock


\bibitem[\protect\citeauthoryear{Ge, He, Ke, and Sun}{Ge et~al\mbox{.}}{2013}]%
        {OPQ}
\bibfield{author}{\bibinfo{person}{Tiezheng Ge}, \bibinfo{person}{Kaiming He}, \bibinfo{person}{Qifa Ke}, {and} \bibinfo{person}{Jian Sun}.} \bibinfo{year}{2013}\natexlab{}.
\newblock \showarticletitle{Optimized product quantization}.
\newblock \bibinfo{journal}{\emph{IEEE transactions on pattern analysis and machine intelligence}} \bibinfo{volume}{36}, \bibinfo{number}{4} (\bibinfo{year}{2013}), \bibinfo{pages}{744--755}.
\newblock


\bibitem[\protect\citeauthoryear{Gionis, Indyk, Motwani, et~al\mbox{.}}{Gionis et~al\mbox{.}}{1999}]%
        {gionis1999similarity}
\bibfield{author}{\bibinfo{person}{Aristides Gionis}, \bibinfo{person}{Piotr Indyk}, \bibinfo{person}{Rajeev Motwani}, {et~al\mbox{.}}} \bibinfo{year}{1999}\natexlab{}.
\newblock \showarticletitle{Similarity search in high dimensions via hashing}. In \bibinfo{booktitle}{\emph{Vldb}}, Vol.~\bibinfo{volume}{99}. \bibinfo{pages}{518--529}.
\newblock


\bibitem[\protect\citeauthoryear{Guo, Luan, Xiang, Yan, Yi, Luo, Cheng, Xu, Luo, Liu, et~al\mbox{.}}{Guo et~al\mbox{.}}{2022}]%
        {guo2022manu}
\bibfield{author}{\bibinfo{person}{Rentong Guo}, \bibinfo{person}{Xiaofan Luan}, \bibinfo{person}{Long Xiang}, \bibinfo{person}{Xiao Yan}, \bibinfo{person}{Xiaomeng Yi}, \bibinfo{person}{Jigao Luo}, \bibinfo{person}{Qianya Cheng}, \bibinfo{person}{Weizhi Xu}, \bibinfo{person}{Jiarui Luo}, \bibinfo{person}{Frank Liu}, {et~al\mbox{.}}} \bibinfo{year}{2022}\natexlab{}.
\newblock \showarticletitle{Manu: A Cloud Native Vector Database Management System}.
\newblock \bibinfo{journal}{\emph{arXiv preprint arXiv:2206.13843}} (\bibinfo{year}{2022}).
\newblock


\bibitem[\protect\citeauthoryear{Guu, Lee, Tung, Pasupat, and Chang}{Guu et~al\mbox{.}}{2020a}]%
        {guu2020retrieval}
\bibfield{author}{\bibinfo{person}{Kelvin Guu}, \bibinfo{person}{Kenton Lee}, \bibinfo{person}{Zora Tung}, \bibinfo{person}{Panupong Pasupat}, {and} \bibinfo{person}{Mingwei Chang}.} \bibinfo{year}{2020}\natexlab{a}.
\newblock \showarticletitle{Retrieval augmented language model pre-training}. In \bibinfo{booktitle}{\emph{International conference on machine learning}}. PMLR, \bibinfo{pages}{3929--3938}.
\newblock


\bibitem[\protect\citeauthoryear{Guu, Lee, Tung, Pasupat, and Chang}{Guu et~al\mbox{.}}{2020b}]%
        {guu2020realm}
\bibfield{author}{\bibinfo{person}{Kelvin Guu}, \bibinfo{person}{Kenton Lee}, \bibinfo{person}{Zora Tung}, \bibinfo{person}{Panupong Pasupat}, {and} \bibinfo{person}{Ming-Wei Chang}.} \bibinfo{year}{2020}\natexlab{b}.
\newblock \showarticletitle{Realm: Retrieval-augmented language model pre-training}.
\newblock \bibinfo{journal}{\emph{arXiv preprint arXiv:2002.08909}} (\bibinfo{year}{2020}).
\newblock


\bibitem[\protect\citeauthoryear{He, Korolija, and Alonso}{He et~al\mbox{.}}{2021}]%
        {100gbps}
\bibfield{author}{\bibinfo{person}{Zhenhao He}, \bibinfo{person}{Dario Korolija}, {and} \bibinfo{person}{Gustavo Alonso}.} \bibinfo{year}{2021}\natexlab{}.
\newblock \showarticletitle{EasyNet: 100 Gbps Network for HLS}. In \bibinfo{booktitle}{\emph{2021 31th International Conference on Field Programmable Logic and Applications (FPL)}}.
\newblock


\bibitem[\protect\citeauthoryear{Hoefler, Lichei, and Rehm}{Hoefler et~al\mbox{.}}{2007}]%
        {hoefler2007low}
\bibfield{author}{\bibinfo{person}{Torsten Hoefler}, \bibinfo{person}{Andre Lichei}, {and} \bibinfo{person}{Wolfgang Rehm}.} \bibinfo{year}{2007}\natexlab{}.
\newblock \showarticletitle{Low-overhead LogGP parameter assessment for modern interconnection networks}. In \bibinfo{booktitle}{\emph{2007 IEEE International Parallel and Distributed Processing Symposium}}. IEEE, \bibinfo{pages}{1--8}.
\newblock


\bibitem[\protect\citeauthoryear{Hoefler and Moor}{Hoefler and Moor}{2014}]%
        {hoefler2014energy}
\bibfield{author}{\bibinfo{person}{Torsten Hoefler} {and} \bibinfo{person}{Dmitry Moor}.} \bibinfo{year}{2014}\natexlab{}.
\newblock \showarticletitle{Energy, memory, and runtime tradeoffs for implementing collective communication operations}.
\newblock \bibinfo{journal}{\emph{Supercomputing frontiers and innovations}} \bibinfo{volume}{1}, \bibinfo{number}{2} (\bibinfo{year}{2014}), \bibinfo{pages}{58--75}.
\newblock


\bibitem[\protect\citeauthoryear{Hu, Wang, Chang, Lee, Lin, Wang, Lin, Huang, Lee, Su, et~al\mbox{.}}{Hu et~al\mbox{.}}{2022}]%
        {hu2022ice}
\bibfield{author}{\bibinfo{person}{Han-Wen Hu}, \bibinfo{person}{Wei-Chen Wang}, \bibinfo{person}{Yuan-Hao Chang}, \bibinfo{person}{Yung-Chun Lee}, \bibinfo{person}{Bo-Rong Lin}, \bibinfo{person}{Huai-Mu Wang}, \bibinfo{person}{Yen-Po Lin}, \bibinfo{person}{Yu-Ming Huang}, \bibinfo{person}{Chong-Ying Lee}, \bibinfo{person}{Tzu-Hsiang Su}, {et~al\mbox{.}}} \bibinfo{year}{2022}\natexlab{}.
\newblock \showarticletitle{ICE: An Intelligent Cognition Engine with 3D NAND-based In-Memory Computing for Vector Similarity Search Acceleration}. In \bibinfo{booktitle}{\emph{2022 55th IEEE/ACM International Symposium on Microarchitecture (MICRO)}}. IEEE, \bibinfo{pages}{763--783}.
\newblock


\bibitem[\protect\citeauthoryear{Huang, Lim, and Cong}{Huang et~al\mbox{.}}{2014}]%
        {huang2014scalable}
\bibfield{author}{\bibinfo{person}{Muhuan Huang}, \bibinfo{person}{Kevin Lim}, {and} \bibinfo{person}{Jason Cong}.} \bibinfo{year}{2014}\natexlab{}.
\newblock \showarticletitle{A scalable, high-performance customized priority queue}. In \bibinfo{booktitle}{\emph{2014 24th International Conference on Field Programmable Logic and Applications (FPL)}}. IEEE, \bibinfo{pages}{1--4}.
\newblock


\bibitem[\protect\citeauthoryear{Huang, Lei, and Tung}{Huang et~al\mbox{.}}{2021}]%
        {huang2021point}
\bibfield{author}{\bibinfo{person}{Qiang Huang}, \bibinfo{person}{Yifan Lei}, {and} \bibinfo{person}{Anthony~KH Tung}.} \bibinfo{year}{2021}\natexlab{}.
\newblock \showarticletitle{Point-to-Hyperplane Nearest Neighbor Search Beyond the Unit Hypersphere}. In \bibinfo{booktitle}{\emph{Proceedings of the 2021 International Conference on Management of Data}}. \bibinfo{pages}{777--789}.
\newblock


\bibitem[\protect\citeauthoryear{Izacard and Grave}{Izacard and Grave}{2020}]%
        {izacard2020leveraging}
\bibfield{author}{\bibinfo{person}{Gautier Izacard} {and} \bibinfo{person}{Edouard Grave}.} \bibinfo{year}{2020}\natexlab{}.
\newblock \showarticletitle{Leveraging passage retrieval with generative models for open domain question answering}.
\newblock \bibinfo{journal}{\emph{arXiv preprint arXiv:2007.01282}} (\bibinfo{year}{2020}).
\newblock


\bibitem[\protect\citeauthoryear{Izacard, Lewis, Lomeli, Hosseini, Petroni, Schick, Dwivedi-Yu, Joulin, Riedel, and Grave}{Izacard et~al\mbox{.}}{2022}]%
        {izacard2022few}
\bibfield{author}{\bibinfo{person}{Gautier Izacard}, \bibinfo{person}{Patrick Lewis}, \bibinfo{person}{Maria Lomeli}, \bibinfo{person}{Lucas Hosseini}, \bibinfo{person}{Fabio Petroni}, \bibinfo{person}{Timo Schick}, \bibinfo{person}{Jane Dwivedi-Yu}, \bibinfo{person}{Armand Joulin}, \bibinfo{person}{Sebastian Riedel}, {and} \bibinfo{person}{Edouard Grave}.} \bibinfo{year}{2022}\natexlab{}.
\newblock \showarticletitle{Few-shot learning with retrieval augmented language models}.
\newblock \bibinfo{journal}{\emph{arXiv preprint arXiv:2208.03299}} (\bibinfo{year}{2022}).
\newblock


\bibitem[\protect\citeauthoryear{Jang, Choi, Bae, Lee, Kwon, and Jung}{Jang et~al\mbox{.}}{2023}]%
        {jang2023cxl}
\bibfield{author}{\bibinfo{person}{Junhyeok Jang}, \bibinfo{person}{Hanjin Choi}, \bibinfo{person}{Hanyeoreum Bae}, \bibinfo{person}{Seungjun Lee}, \bibinfo{person}{Miryeong Kwon}, {and} \bibinfo{person}{Myoungsoo Jung}.} \bibinfo{year}{2023}\natexlab{}.
\newblock \showarticletitle{$\{$CXL-ANNS$\}$:$\{$Software-Hardware$\}$ Collaborative Memory Disaggregation and Computation for $\{$Billion-Scale$\}$ Approximate Nearest Neighbor Search}. In \bibinfo{booktitle}{\emph{2023 USENIX Annual Technical Conference (USENIX ATC 23)}}. \bibinfo{pages}{585--600}.
\newblock


\bibitem[\protect\citeauthoryear{Jayaram~Subramanya, Devvrit, Simhadri, Krishnawamy, and Kadekodi}{Jayaram~Subramanya et~al\mbox{.}}{2019}]%
        {jayaram2019diskann}
\bibfield{author}{\bibinfo{person}{Suhas Jayaram~Subramanya}, \bibinfo{person}{Fnu Devvrit}, \bibinfo{person}{Harsha~Vardhan Simhadri}, \bibinfo{person}{Ravishankar Krishnawamy}, {and} \bibinfo{person}{Rohan Kadekodi}.} \bibinfo{year}{2019}\natexlab{}.
\newblock \showarticletitle{Diskann: Fast accurate billion-point nearest neighbor search on a single node}.
\newblock \bibinfo{journal}{\emph{Advances in Neural Information Processing Systems}}  \bibinfo{volume}{32} (\bibinfo{year}{2019}).
\newblock


\bibitem[\protect\citeauthoryear{Jegou, Douze, and Schmid}{Jegou et~al\mbox{.}}{2010}]%
        {PQ}
\bibfield{author}{\bibinfo{person}{Herve Jegou}, \bibinfo{person}{Matthijs Douze}, {and} \bibinfo{person}{Cordelia Schmid}.} \bibinfo{year}{2010}\natexlab{}.
\newblock \showarticletitle{Product quantization for nearest neighbor search}.
\newblock \bibinfo{journal}{\emph{IEEE transactions on pattern analysis and machine intelligence}} \bibinfo{volume}{33}, \bibinfo{number}{1} (\bibinfo{year}{2010}), \bibinfo{pages}{117--128}.
\newblock


\bibitem[\protect\citeauthoryear{Jiang, Fu, and Wong}{Jiang et~al\mbox{.}}{2015}]%
        {jiang2015exact}
\bibfield{author}{\bibinfo{person}{Minhao Jiang}, \bibinfo{person}{Ada Wai-Chee Fu}, {and} \bibinfo{person}{Raymond Chi-Wing Wong}.} \bibinfo{year}{2015}\natexlab{}.
\newblock \showarticletitle{Exact top-k nearest keyword search in large networks}. In \bibinfo{booktitle}{\emph{Proceedings of the 2015 ACM SIGMOD international conference on management of data}}. \bibinfo{pages}{393--404}.
\newblock


\bibitem[\protect\citeauthoryear{Jiang, Hu, Hoefler, and Alonso}{Jiang et~al\mbox{.}}{2024a}]%
        {jiang2024accelerating}
\bibfield{author}{\bibinfo{person}{Wenqi Jiang}, \bibinfo{person}{Hang Hu}, \bibinfo{person}{Torsten Hoefler}, {and} \bibinfo{person}{Gustavo Alonso}.} \bibinfo{year}{2024}\natexlab{a}.
\newblock \showarticletitle{Accelerating Graph-based Vector Search via Delayed-Synchronization Traversal}.
\newblock \bibinfo{journal}{\emph{arXiv preprint arXiv:2406.12385}} (\bibinfo{year}{2024}).
\newblock


\bibitem[\protect\citeauthoryear{Jiang, Li, Zhu, Licht, He, Shi, Renggli, Zhang, Rekatsinas, Hoefler, et~al\mbox{.}}{Jiang et~al\mbox{.}}{2023}]%
        {jiang2023co}
\bibfield{author}{\bibinfo{person}{Wenqi Jiang}, \bibinfo{person}{Shigang Li}, \bibinfo{person}{Yu Zhu}, \bibinfo{person}{Johannes de~Fine Licht}, \bibinfo{person}{Zhenhao He}, \bibinfo{person}{Runbin Shi}, \bibinfo{person}{Cedric Renggli}, \bibinfo{person}{Shuai Zhang}, \bibinfo{person}{Theodoros Rekatsinas}, \bibinfo{person}{Torsten Hoefler}, {et~al\mbox{.}}} \bibinfo{year}{2023}\natexlab{}.
\newblock \showarticletitle{Co-design Hardware and Algorithm for Vector Search}.
\newblock \bibinfo{journal}{\emph{arXiv preprint arXiv:2306.11182}} (\bibinfo{year}{2023}).
\newblock


\bibitem[\protect\citeauthoryear{Jiang, Zhang, Han, Wang, Wang, and Kraska}{Jiang et~al\mbox{.}}{2024b}]%
        {jiang2024piperag}
\bibfield{author}{\bibinfo{person}{Wenqi Jiang}, \bibinfo{person}{Shuai Zhang}, \bibinfo{person}{Boran Han}, \bibinfo{person}{Jie Wang}, \bibinfo{person}{Bernie Wang}, {and} \bibinfo{person}{Tim Kraska}.} \bibinfo{year}{2024}\natexlab{b}.
\newblock \showarticletitle{Piperag: Fast retrieval-augmented generation via algorithm-system co-design}.
\newblock \bibinfo{journal}{\emph{arXiv preprint arXiv:2403.05676}} (\bibinfo{year}{2024}).
\newblock


\bibitem[\protect\citeauthoryear{Johnson, Douze, and J{\'e}gou}{Johnson et~al\mbox{.}}{2019}]%
        {johnson2019billion}
\bibfield{author}{\bibinfo{person}{Jeff Johnson}, \bibinfo{person}{Matthijs Douze}, {and} \bibinfo{person}{Herv{\'e} J{\'e}gou}.} \bibinfo{year}{2019}\natexlab{}.
\newblock \showarticletitle{Billion-scale similarity search with gpus}.
\newblock \bibinfo{journal}{\emph{IEEE Transactions on Big Data}} (\bibinfo{year}{2019}).
\newblock


\bibitem[\protect\citeauthoryear{Khandelwal, Fan, Jurafsky, Zettlemoyer, and Lewis}{Khandelwal et~al\mbox{.}}{2020}]%
        {khandelwal2020nearest}
\bibfield{author}{\bibinfo{person}{Urvashi Khandelwal}, \bibinfo{person}{Angela Fan}, \bibinfo{person}{Dan Jurafsky}, \bibinfo{person}{Luke Zettlemoyer}, {and} \bibinfo{person}{Mike Lewis}.} \bibinfo{year}{2020}\natexlab{}.
\newblock \showarticletitle{Nearest neighbor machine translation}.
\newblock \bibinfo{journal}{\emph{arXiv preprint arXiv:2010.00710}} (\bibinfo{year}{2020}).
\newblock


\bibitem[\protect\citeauthoryear{Khandelwal, Levy, Jurafsky, Zettlemoyer, and Lewis}{Khandelwal et~al\mbox{.}}{2019}]%
        {khandelwal2019generalization}
\bibfield{author}{\bibinfo{person}{Urvashi Khandelwal}, \bibinfo{person}{Omer Levy}, \bibinfo{person}{Dan Jurafsky}, \bibinfo{person}{Luke Zettlemoyer}, {and} \bibinfo{person}{Mike Lewis}.} \bibinfo{year}{2019}\natexlab{}.
\newblock \showarticletitle{Generalization through memorization: Nearest neighbor language models}.
\newblock \bibinfo{journal}{\emph{arXiv preprint arXiv:1911.00172}} (\bibinfo{year}{2019}).
\newblock


\bibitem[\protect\citeauthoryear{Komeili, Shuster, and Weston}{Komeili et~al\mbox{.}}{2021}]%
        {komeili2021internet}
\bibfield{author}{\bibinfo{person}{Mojtaba Komeili}, \bibinfo{person}{Kurt Shuster}, {and} \bibinfo{person}{Jason Weston}.} \bibinfo{year}{2021}\natexlab{}.
\newblock \showarticletitle{Internet-augmented dialogue generation}.
\newblock \bibinfo{journal}{\emph{arXiv preprint arXiv:2107.07566}} (\bibinfo{year}{2021}).
\newblock


\bibitem[\protect\citeauthoryear{Kwon, Li, Zhuang, Sheng, Zheng, Yu, Gonzalez, Zhang, and Stoica}{Kwon et~al\mbox{.}}{2023}]%
        {kwon2023efficient}
\bibfield{author}{\bibinfo{person}{Woosuk Kwon}, \bibinfo{person}{Zhuohan Li}, \bibinfo{person}{Siyuan Zhuang}, \bibinfo{person}{Ying Sheng}, \bibinfo{person}{Lianmin Zheng}, \bibinfo{person}{Cody~Hao Yu}, \bibinfo{person}{Joseph~E. Gonzalez}, \bibinfo{person}{Hao Zhang}, {and} \bibinfo{person}{Ion Stoica}.} \bibinfo{year}{2023}\natexlab{}.
\newblock \showarticletitle{Efficient Memory Management for Large Language Model Serving with PagedAttention}. In \bibinfo{booktitle}{\emph{Proceedings of the ACM SIGOPS 29th Symposium on Operating Systems Principles}}.
\newblock


\bibitem[\protect\citeauthoryear{Lee, Choi, Min, Lee, Beak, Jeong, Lee, and Ham}{Lee et~al\mbox{.}}{2022}]%
        {lee2022anna}
\bibfield{author}{\bibinfo{person}{Yejin Lee}, \bibinfo{person}{Hyunji Choi}, \bibinfo{person}{Sunhong Min}, \bibinfo{person}{Hyunseung Lee}, \bibinfo{person}{Sangwon Beak}, \bibinfo{person}{Dawoon Jeong}, \bibinfo{person}{Jae~W Lee}, {and} \bibinfo{person}{Tae~Jun Ham}.} \bibinfo{year}{2022}\natexlab{}.
\newblock \showarticletitle{ANNA: Specialized Architecture for Approximate Nearest Neighbor Search}. In \bibinfo{booktitle}{\emph{2022 IEEE International Symposium on High-Performance Computer Architecture (HPCA)}}. IEEE, \bibinfo{pages}{169--183}.
\newblock


\bibitem[\protect\citeauthoryear{Leiserson}{Leiserson}{1979}]%
        {leiserson1979systolic}
\bibfield{author}{\bibinfo{person}{Charles~E Leiserson}.} \bibinfo{year}{1979}\natexlab{}.
\newblock \bibinfo{booktitle}{\emph{Systolic Priority Queues.}}
\newblock \bibinfo{type}{{T}echnical {R}eport}. \bibinfo{institution}{CARNEGIE-MELLON UNIV PITTSBURGH PA DEPT OF COMPUTER SCIENCE}.
\newblock


\bibitem[\protect\citeauthoryear{Lejsek, {\'A}smundsson, J{\'o}nsson, and Amsaleg}{Lejsek et~al\mbox{.}}{2008}]%
        {lejsek2008nv}
\bibfield{author}{\bibinfo{person}{Herwig Lejsek}, \bibinfo{person}{Fri{\dh}rik~Hei{\dh}ar {\'A}smundsson}, \bibinfo{person}{Bj{\"o}rn~{\TH}{\'o}r J{\'o}nsson}, {and} \bibinfo{person}{Laurent Amsaleg}.} \bibinfo{year}{2008}\natexlab{}.
\newblock \showarticletitle{NV-Tree: An efficient disk-based index for approximate search in very large high-dimensional collections}.
\newblock \bibinfo{journal}{\emph{IEEE Transactions on Pattern Analysis and Machine Intelligence}} \bibinfo{volume}{31}, \bibinfo{number}{5} (\bibinfo{year}{2008}), \bibinfo{pages}{869--883}.
\newblock


\bibitem[\protect\citeauthoryear{Lewis, Ghazvininejad, Ghosh, Aghajanyan, Wang, and Zettlemoyer}{Lewis et~al\mbox{.}}{2020a}]%
        {lewis2020pre}
\bibfield{author}{\bibinfo{person}{Mike Lewis}, \bibinfo{person}{Marjan Ghazvininejad}, \bibinfo{person}{Gargi Ghosh}, \bibinfo{person}{Armen Aghajanyan}, \bibinfo{person}{Sida Wang}, {and} \bibinfo{person}{Luke Zettlemoyer}.} \bibinfo{year}{2020}\natexlab{a}.
\newblock \showarticletitle{Pre-training via paraphrasing}.
\newblock \bibinfo{journal}{\emph{Advances in Neural Information Processing Systems}}  \bibinfo{volume}{33} (\bibinfo{year}{2020}), \bibinfo{pages}{18470--18481}.
\newblock


\bibitem[\protect\citeauthoryear{Lewis, Perez, Piktus, Petroni, Karpukhin, Goyal, K{\"u}ttler, Lewis, Yih, Rockt{\"a}schel, et~al\mbox{.}}{Lewis et~al\mbox{.}}{2020b}]%
        {lewis2020retrieval}
\bibfield{author}{\bibinfo{person}{Patrick Lewis}, \bibinfo{person}{Ethan Perez}, \bibinfo{person}{Aleksandra Piktus}, \bibinfo{person}{Fabio Petroni}, \bibinfo{person}{Vladimir Karpukhin}, \bibinfo{person}{Naman Goyal}, \bibinfo{person}{Heinrich K{\"u}ttler}, \bibinfo{person}{Mike Lewis}, \bibinfo{person}{Wen-tau Yih}, \bibinfo{person}{Tim Rockt{\"a}schel}, {et~al\mbox{.}}} \bibinfo{year}{2020}\natexlab{b}.
\newblock \showarticletitle{Retrieval-augmented generation for knowledge-intensive nlp tasks}.
\newblock \bibinfo{journal}{\emph{Advances in Neural Information Processing Systems}}  \bibinfo{volume}{33} (\bibinfo{year}{2020}), \bibinfo{pages}{9459--9474}.
\newblock


\bibitem[\protect\citeauthoryear{Li}{Li}{2023}]%
        {li2023dark}
\bibfield{author}{\bibinfo{person}{Zihao Li}.} \bibinfo{year}{2023}\natexlab{}.
\newblock \showarticletitle{The dark side of chatgpt: Legal and ethical challenges from stochastic parrots and hallucination}.
\newblock \bibinfo{journal}{\emph{arXiv preprint arXiv:2304.14347}} (\bibinfo{year}{2023}).
\newblock


\bibitem[\protect\citeauthoryear{Li, Guo, and Kumar}{Li et~al\mbox{.}}{2022}]%
        {li2022decoupled}
\bibfield{author}{\bibinfo{person}{Zonglin Li}, \bibinfo{person}{Ruiqi Guo}, {and} \bibinfo{person}{Sanjiv Kumar}.} \bibinfo{year}{2022}\natexlab{}.
\newblock \showarticletitle{Decoupled context processing for context augmented language modeling}.
\newblock \bibinfo{journal}{\emph{Advances in Neural Information Processing Systems}}  \bibinfo{volume}{35} (\bibinfo{year}{2022}), \bibinfo{pages}{21698--21710}.
\newblock


\bibitem[\protect\citeauthoryear{Liu, Ni, Leng, Feng, Guo, Chen, Li, Guo, and Zhu}{Liu et~al\mbox{.}}{2023}]%
        {liu2023juno}
\bibfield{author}{\bibinfo{person}{Zihan Liu}, \bibinfo{person}{Wentao Ni}, \bibinfo{person}{Jingwen Leng}, \bibinfo{person}{Yu Feng}, \bibinfo{person}{Cong Guo}, \bibinfo{person}{Quan Chen}, \bibinfo{person}{Chao Li}, \bibinfo{person}{Minyi Guo}, {and} \bibinfo{person}{Yuhao Zhu}.} \bibinfo{year}{2023}\natexlab{}.
\newblock \showarticletitle{JUNO: Optimizing High-Dimensional Approximate Nearest Neighbour Search with Sparsity-Aware Algorithm and Ray-Tracing Core Mapping}.
\newblock \bibinfo{journal}{\emph{arXiv preprint arXiv:2312.01712}} (\bibinfo{year}{2023}).
\newblock


\bibitem[\protect\citeauthoryear{Lu, Kudo, Xiao, and Ishikawa}{Lu et~al\mbox{.}}{2021}]%
        {lu2021hvs}
\bibfield{author}{\bibinfo{person}{Kejing Lu}, \bibinfo{person}{Mineichi Kudo}, \bibinfo{person}{Chuan Xiao}, {and} \bibinfo{person}{Yoshiharu Ishikawa}.} \bibinfo{year}{2021}\natexlab{}.
\newblock \showarticletitle{HVS: hierarchical graph structure based on voronoi diagrams for solving approximate nearest neighbor search}.
\newblock \bibinfo{journal}{\emph{Proceedings of the VLDB Endowment}} \bibinfo{volume}{15}, \bibinfo{number}{2} (\bibinfo{year}{2021}), \bibinfo{pages}{246--258}.
\newblock


\bibitem[\protect\citeauthoryear{Lu, Shen, Chen, and Ooi}{Lu et~al\mbox{.}}{2012}]%
        {lu2012efficient}
\bibfield{author}{\bibinfo{person}{Wei Lu}, \bibinfo{person}{Yanyan Shen}, \bibinfo{person}{Su Chen}, {and} \bibinfo{person}{Beng~Chin Ooi}.} \bibinfo{year}{2012}\natexlab{}.
\newblock \showarticletitle{Efficient processing of k nearest neighbor joins using mapreduce}.
\newblock \bibinfo{journal}{\emph{arXiv preprint arXiv:1207.0141}} (\bibinfo{year}{2012}).
\newblock


\bibitem[\protect\citeauthoryear{Malkov, Ponomarenko, Logvinov, and Krylov}{Malkov et~al\mbox{.}}{2014}]%
        {malkov2014approximate}
\bibfield{author}{\bibinfo{person}{Yury Malkov}, \bibinfo{person}{Alexander Ponomarenko}, \bibinfo{person}{Andrey Logvinov}, {and} \bibinfo{person}{Vladimir Krylov}.} \bibinfo{year}{2014}\natexlab{}.
\newblock \showarticletitle{Approximate nearest neighbor algorithm based on navigable small world graphs}.
\newblock \bibinfo{journal}{\emph{Information Systems}}  \bibinfo{volume}{45} (\bibinfo{year}{2014}), \bibinfo{pages}{61--68}.
\newblock


\bibitem[\protect\citeauthoryear{Malkov and Yashunin}{Malkov and Yashunin}{2018}]%
        {malkov2018efficient}
\bibfield{author}{\bibinfo{person}{Yu~A Malkov} {and} \bibinfo{person}{Dmitry~A Yashunin}.} \bibinfo{year}{2018}\natexlab{}.
\newblock \showarticletitle{Efficient and robust approximate nearest neighbor search using hierarchical navigable small world graphs}.
\newblock \bibinfo{journal}{\emph{IEEE transactions on pattern analysis and machine intelligence}} \bibinfo{volume}{42}, \bibinfo{number}{4} (\bibinfo{year}{2018}), \bibinfo{pages}{824--836}.
\newblock


\bibitem[\protect\citeauthoryear{Meng, Li, Zheng, Wu, Sun, Zhang, and Li}{Meng et~al\mbox{.}}{2021}]%
        {meng2021fast}
\bibfield{author}{\bibinfo{person}{Yuxian Meng}, \bibinfo{person}{Xiaoya Li}, \bibinfo{person}{Xiayu Zheng}, \bibinfo{person}{Fei Wu}, \bibinfo{person}{Xiaofei Sun}, \bibinfo{person}{Tianwei Zhang}, {and} \bibinfo{person}{Jiwei Li}.} \bibinfo{year}{2021}\natexlab{}.
\newblock \showarticletitle{Fast nearest neighbor machine translation}.
\newblock \bibinfo{journal}{\emph{arXiv preprint arXiv:2105.14528}} (\bibinfo{year}{2021}).
\newblock


\bibitem[\protect\citeauthoryear{Mohoney, Pacaci, Chowdhury, Mousavi, Ilyas, Minhas, Pound, and Rekatsinas}{Mohoney et~al\mbox{.}}{2023}]%
        {mohoney2023high}
\bibfield{author}{\bibinfo{person}{Jason Mohoney}, \bibinfo{person}{Anil Pacaci}, \bibinfo{person}{Shihabur~Rahman Chowdhury}, \bibinfo{person}{Ali Mousavi}, \bibinfo{person}{Ihab~F Ilyas}, \bibinfo{person}{Umar~Farooq Minhas}, \bibinfo{person}{Jeffrey Pound}, {and} \bibinfo{person}{Theodoros Rekatsinas}.} \bibinfo{year}{2023}\natexlab{}.
\newblock \showarticletitle{High-Throughput Vector Similarity Search in Knowledge Graphs}.
\newblock \bibinfo{journal}{\emph{Proceedings of the ACM on Management of Data}} \bibinfo{volume}{1}, \bibinfo{number}{2} (\bibinfo{year}{2023}), \bibinfo{pages}{1--25}.
\newblock


\bibitem[\protect\citeauthoryear{Narayanan, Harlap, Phanishayee, Seshadri, Devanur, Ganger, Gibbons, and Zaharia}{Narayanan et~al\mbox{.}}{2019}]%
        {narayanan2019pipedream}
\bibfield{author}{\bibinfo{person}{Deepak Narayanan}, \bibinfo{person}{Aaron Harlap}, \bibinfo{person}{Amar Phanishayee}, \bibinfo{person}{Vivek Seshadri}, \bibinfo{person}{Nikhil~R Devanur}, \bibinfo{person}{Gregory~R Ganger}, \bibinfo{person}{Phillip~B Gibbons}, {and} \bibinfo{person}{Matei Zaharia}.} \bibinfo{year}{2019}\natexlab{}.
\newblock \showarticletitle{PipeDream: generalized pipeline parallelism for DNN training}. In \bibinfo{booktitle}{\emph{Proceedings of the 27th ACM symposium on operating systems principles}}. \bibinfo{pages}{1--15}.
\newblock


\bibitem[\protect\citeauthoryear{Ott, Edunov, Baevski, Fan, Gross, Ng, Grangier, and Auli}{Ott et~al\mbox{.}}{2019}]%
        {ott2019fairseq}
\bibfield{author}{\bibinfo{person}{Myle Ott}, \bibinfo{person}{Sergey Edunov}, \bibinfo{person}{Alexei Baevski}, \bibinfo{person}{Angela Fan}, \bibinfo{person}{Sam Gross}, \bibinfo{person}{Nathan Ng}, \bibinfo{person}{David Grangier}, {and} \bibinfo{person}{Michael Auli}.} \bibinfo{year}{2019}\natexlab{}.
\newblock \showarticletitle{fairseq: A fast, extensible toolkit for sequence modeling}.
\newblock \bibinfo{journal}{\emph{arXiv preprint arXiv:1904.01038}} (\bibinfo{year}{2019}).
\newblock


\bibitem[\protect\citeauthoryear{Ouyang, Wen, Qin, Chang, Zhang, and Lin}{Ouyang et~al\mbox{.}}{2020}]%
        {ouyang2020progressive}
\bibfield{author}{\bibinfo{person}{Dian Ouyang}, \bibinfo{person}{Dong Wen}, \bibinfo{person}{Lu Qin}, \bibinfo{person}{Lijun Chang}, \bibinfo{person}{Ying Zhang}, {and} \bibinfo{person}{Xuemin Lin}.} \bibinfo{year}{2020}\natexlab{}.
\newblock \showarticletitle{Progressive top-k nearest neighbors search in large road networks}. In \bibinfo{booktitle}{\emph{Proceedings of the 2020 ACM SIGMOD International Conference on Management of Data}}. \bibinfo{pages}{1781--1795}.
\newblock


\bibitem[\protect\citeauthoryear{Pan, Wang, and Li}{Pan et~al\mbox{.}}{2023}]%
        {pan2023survey}
\bibfield{author}{\bibinfo{person}{James~Jie Pan}, \bibinfo{person}{Jianguo Wang}, {and} \bibinfo{person}{Guoliang Li}.} \bibinfo{year}{2023}\natexlab{}.
\newblock \showarticletitle{Survey of vector database management systems}.
\newblock \bibinfo{journal}{\emph{arXiv preprint arXiv:2310.14021}} (\bibinfo{year}{2023}).
\newblock


\bibitem[\protect\citeauthoryear{Patel, Choukse, Zhang, Goiri, Shah, Maleki, and Bianchini}{Patel et~al\mbox{.}}{2023}]%
        {patel2023splitwise}
\bibfield{author}{\bibinfo{person}{Pratyush Patel}, \bibinfo{person}{Esha Choukse}, \bibinfo{person}{Chaojie Zhang}, \bibinfo{person}{{\'I}{\~n}igo Goiri}, \bibinfo{person}{Aashaka Shah}, \bibinfo{person}{Saeed Maleki}, {and} \bibinfo{person}{Ricardo Bianchini}.} \bibinfo{year}{2023}\natexlab{}.
\newblock \showarticletitle{Splitwise: Efficient generative llm inference using phase splitting}.
\newblock \bibinfo{journal}{\emph{arXiv preprint arXiv:2311.18677}} (\bibinfo{year}{2023}).
\newblock


\bibitem[\protect\citeauthoryear{Peng, Chen, Wang, Yang, Weitze, Geng, Li, Bi, Song, Jiang, et~al\mbox{.}}{Peng et~al\mbox{.}}{2021}]%
        {peng2021optimizing}
\bibfield{author}{\bibinfo{person}{Hongwu Peng}, \bibinfo{person}{Shiyang Chen}, \bibinfo{person}{Zhepeng Wang}, \bibinfo{person}{Junhuan Yang}, \bibinfo{person}{Scott~A Weitze}, \bibinfo{person}{Tong Geng}, \bibinfo{person}{Ang Li}, \bibinfo{person}{Jinbo Bi}, \bibinfo{person}{Minghu Song}, \bibinfo{person}{Weiwen Jiang}, {et~al\mbox{.}}} \bibinfo{year}{2021}\natexlab{}.
\newblock \showarticletitle{Optimizing fpga-based accelerator design for large-scale molecular similarity search (special session paper)}. In \bibinfo{booktitle}{\emph{2021 IEEE/ACM International Conference On Computer Aided Design (ICCAD)}}. IEEE, \bibinfo{pages}{1--7}.
\newblock


\bibitem[\protect\citeauthoryear{Peng, Choi, Chan, Yang, and Xu}{Peng et~al\mbox{.}}{2023}]%
        {peng2023efficient}
\bibfield{author}{\bibinfo{person}{Yun Peng}, \bibinfo{person}{Byron Choi}, \bibinfo{person}{Tsz~Nam Chan}, \bibinfo{person}{Jianye Yang}, {and} \bibinfo{person}{Jianliang Xu}.} \bibinfo{year}{2023}\natexlab{}.
\newblock \showarticletitle{Efficient approximate nearest neighbor search in multi-dimensional databases}.
\newblock \bibinfo{journal}{\emph{Proceedings of the ACM on Management of Data}} \bibinfo{volume}{1}, \bibinfo{number}{1} (\bibinfo{year}{2023}), \bibinfo{pages}{1--27}.
\newblock


\bibitem[\protect\citeauthoryear{Radford, Wu, Child, Luan, Amodei, Sutskever, et~al\mbox{.}}{Radford et~al\mbox{.}}{2019}]%
        {radford2019language}
\bibfield{author}{\bibinfo{person}{Alec Radford}, \bibinfo{person}{Jeffrey Wu}, \bibinfo{person}{Rewon Child}, \bibinfo{person}{David Luan}, \bibinfo{person}{Dario Amodei}, \bibinfo{person}{Ilya Sutskever}, {et~al\mbox{.}}} \bibinfo{year}{2019}\natexlab{}.
\newblock \showarticletitle{Language models are unsupervised multitask learners}.
\newblock \bibinfo{journal}{\emph{OpenAI blog}} \bibinfo{volume}{1}, \bibinfo{number}{8} (\bibinfo{year}{2019}), \bibinfo{pages}{9}.
\newblock


\bibitem[\protect\citeauthoryear{Rae, Borgeaud, Cai, Millican, Hoffmann, Song, Aslanides, Henderson, Ring, Young, et~al\mbox{.}}{Rae et~al\mbox{.}}{2021}]%
        {rae2021scaling}
\bibfield{author}{\bibinfo{person}{Jack~W Rae}, \bibinfo{person}{Sebastian Borgeaud}, \bibinfo{person}{Trevor Cai}, \bibinfo{person}{Katie Millican}, \bibinfo{person}{Jordan Hoffmann}, \bibinfo{person}{Francis Song}, \bibinfo{person}{John Aslanides}, \bibinfo{person}{Sarah Henderson}, \bibinfo{person}{Roman Ring}, \bibinfo{person}{Susannah Young}, {et~al\mbox{.}}} \bibinfo{year}{2021}\natexlab{}.
\newblock \showarticletitle{Scaling language models: Methods, analysis \& insights from training gopher}.
\newblock \bibinfo{journal}{\emph{arXiv preprint arXiv:2112.11446}} (\bibinfo{year}{2021}).
\newblock


\bibitem[\protect\citeauthoryear{Raffel, Shazeer, Roberts, Lee, Narang, Matena, Zhou, Li, and Liu}{Raffel et~al\mbox{.}}{2020}]%
        {raffel2020exploring}
\bibfield{author}{\bibinfo{person}{Colin Raffel}, \bibinfo{person}{Noam Shazeer}, \bibinfo{person}{Adam Roberts}, \bibinfo{person}{Katherine Lee}, \bibinfo{person}{Sharan Narang}, \bibinfo{person}{Michael Matena}, \bibinfo{person}{Yanqi Zhou}, \bibinfo{person}{Wei Li}, {and} \bibinfo{person}{Peter~J Liu}.} \bibinfo{year}{2020}\natexlab{}.
\newblock \showarticletitle{Exploring the limits of transfer learning with a unified text-to-text transformer}.
\newblock \bibinfo{journal}{\emph{The Journal of Machine Learning Research}} \bibinfo{volume}{21}, \bibinfo{number}{1} (\bibinfo{year}{2020}), \bibinfo{pages}{5485--5551}.
\newblock


\bibitem[\protect\citeauthoryear{Rajbhandari, Rasley, Ruwase, and He}{Rajbhandari et~al\mbox{.}}{2020}]%
        {rajbhandari2020zero}
\bibfield{author}{\bibinfo{person}{Samyam Rajbhandari}, \bibinfo{person}{Jeff Rasley}, \bibinfo{person}{Olatunji Ruwase}, {and} \bibinfo{person}{Yuxiong He}.} \bibinfo{year}{2020}\natexlab{}.
\newblock \showarticletitle{Zero: Memory optimizations toward training trillion parameter models}. In \bibinfo{booktitle}{\emph{SC20: International Conference for High Performance Computing, Networking, Storage and Analysis}}. IEEE, \bibinfo{pages}{1--16}.
\newblock


\bibitem[\protect\citeauthoryear{Ram, Levine, Dalmedigos, Muhlgay, Shashua, Leyton-Brown, and Shoham}{Ram et~al\mbox{.}}{2023}]%
        {ram2023context}
\bibfield{author}{\bibinfo{person}{Ori Ram}, \bibinfo{person}{Yoav Levine}, \bibinfo{person}{Itay Dalmedigos}, \bibinfo{person}{Dor Muhlgay}, \bibinfo{person}{Amnon Shashua}, \bibinfo{person}{Kevin Leyton-Brown}, {and} \bibinfo{person}{Yoav Shoham}.} \bibinfo{year}{2023}\natexlab{}.
\newblock \showarticletitle{In-context retrieval-augmented language models}.
\newblock \bibinfo{journal}{\emph{arXiv preprint arXiv:2302.00083}} (\bibinfo{year}{2023}).
\newblock


\bibitem[\protect\citeauthoryear{Ren, Zhang, and Li}{Ren et~al\mbox{.}}{2020}]%
        {ren2020hm}
\bibfield{author}{\bibinfo{person}{Jie Ren}, \bibinfo{person}{Minjia Zhang}, {and} \bibinfo{person}{Dong Li}.} \bibinfo{year}{2020}\natexlab{}.
\newblock \showarticletitle{Hm-ann: Efficient billion-point nearest neighbor search on heterogeneous memory}.
\newblock \bibinfo{journal}{\emph{Advances in Neural Information Processing Systems}}  \bibinfo{volume}{33} (\bibinfo{year}{2020}), \bibinfo{pages}{10672--10684}.
\newblock


\bibitem[\protect\citeauthoryear{Sachan, Patwary, Shoeybi, Kant, Ping, Hamilton, and Catanzaro}{Sachan et~al\mbox{.}}{2021}]%
        {sachan2021end}
\bibfield{author}{\bibinfo{person}{Devendra~Singh Sachan}, \bibinfo{person}{Mostofa Patwary}, \bibinfo{person}{Mohammad Shoeybi}, \bibinfo{person}{Neel Kant}, \bibinfo{person}{Wei Ping}, \bibinfo{person}{William~L Hamilton}, {and} \bibinfo{person}{Bryan Catanzaro}.} \bibinfo{year}{2021}\natexlab{}.
\newblock \showarticletitle{End-to-end training of neural retrievers for open-domain question answering}.
\newblock \bibinfo{journal}{\emph{arXiv preprint arXiv:2101.00408}} (\bibinfo{year}{2021}).
\newblock


\bibitem[\protect\citeauthoryear{Shoeybi, Patwary, Puri, LeGresley, Casper, and Catanzaro}{Shoeybi et~al\mbox{.}}{2019}]%
        {shoeybi2019megatron}
\bibfield{author}{\bibinfo{person}{Mohammad Shoeybi}, \bibinfo{person}{Mostofa Patwary}, \bibinfo{person}{Raul Puri}, \bibinfo{person}{Patrick LeGresley}, \bibinfo{person}{Jared Casper}, {and} \bibinfo{person}{Bryan Catanzaro}.} \bibinfo{year}{2019}\natexlab{}.
\newblock \showarticletitle{Megatron-lm: Training multi-billion parameter language models using model parallelism}.
\newblock \bibinfo{journal}{\emph{arXiv preprint arXiv:1909.08053}} (\bibinfo{year}{2019}).
\newblock


\bibitem[\protect\citeauthoryear{Smith, Patwary, Norick, LeGresley, Rajbhandari, Casper, Liu, Prabhumoye, Zerveas, Korthikanti, et~al\mbox{.}}{Smith et~al\mbox{.}}{2022}]%
        {smith2022using}
\bibfield{author}{\bibinfo{person}{Shaden Smith}, \bibinfo{person}{Mostofa Patwary}, \bibinfo{person}{Brandon Norick}, \bibinfo{person}{Patrick LeGresley}, \bibinfo{person}{Samyam Rajbhandari}, \bibinfo{person}{Jared Casper}, \bibinfo{person}{Zhun Liu}, \bibinfo{person}{Shrimai Prabhumoye}, \bibinfo{person}{George Zerveas}, \bibinfo{person}{Vijay Korthikanti}, {et~al\mbox{.}}} \bibinfo{year}{2022}\natexlab{}.
\newblock \showarticletitle{Using deepspeed and megatron to train megatron-turing nlg 530b, a large-scale generative language model}.
\newblock \bibinfo{journal}{\emph{arXiv preprint arXiv:2201.11990}} (\bibinfo{year}{2022}).
\newblock


\bibitem[\protect\citeauthoryear{Sun, Wang, Qin, Zhang, and Lin}{Sun et~al\mbox{.}}{2014}]%
        {sun2014srs}
\bibfield{author}{\bibinfo{person}{Yifang Sun}, \bibinfo{person}{Wei Wang}, \bibinfo{person}{Jianbin Qin}, \bibinfo{person}{Ying Zhang}, {and} \bibinfo{person}{Xuemin Lin}.} \bibinfo{year}{2014}\natexlab{}.
\newblock \showarticletitle{SRS: solving c-approximate nearest neighbor queries in high dimensional euclidean space with a tiny index}.
\newblock \bibinfo{journal}{\emph{Proceedings of the VLDB Endowment}} (\bibinfo{year}{2014}).
\newblock


\bibitem[\protect\citeauthoryear{Wang, Ping, McAfee, Xu, Li, Shoeybi, and Catanzaro}{Wang et~al\mbox{.}}{2023}]%
        {wang2023instructretro}
\bibfield{author}{\bibinfo{person}{Boxin Wang}, \bibinfo{person}{Wei Ping}, \bibinfo{person}{Lawrence McAfee}, \bibinfo{person}{Peng Xu}, \bibinfo{person}{Bo Li}, \bibinfo{person}{Mohammad Shoeybi}, {and} \bibinfo{person}{Bryan Catanzaro}.} \bibinfo{year}{2023}\natexlab{}.
\newblock \showarticletitle{Instructretro: Instruction tuning post retrieval-augmented pretraining}.
\newblock \bibinfo{journal}{\emph{arXiv preprint arXiv:2310.07713}} (\bibinfo{year}{2023}).
\newblock


\bibitem[\protect\citeauthoryear{Wang, Yi, Guo, Jin, Xu, Li, Wang, Guo, Li, Xu, et~al\mbox{.}}{Wang et~al\mbox{.}}{2021}]%
        {wang2021milvus}
\bibfield{author}{\bibinfo{person}{Jianguo Wang}, \bibinfo{person}{Xiaomeng Yi}, \bibinfo{person}{Rentong Guo}, \bibinfo{person}{Hai Jin}, \bibinfo{person}{Peng Xu}, \bibinfo{person}{Shengjun Li}, \bibinfo{person}{Xiangyu Wang}, \bibinfo{person}{Xiangzhou Guo}, \bibinfo{person}{Chengming Li}, \bibinfo{person}{Xiaohai Xu}, {et~al\mbox{.}}} \bibinfo{year}{2021}\natexlab{}.
\newblock \showarticletitle{Milvus: A purpose-built vector data management system}. In \bibinfo{booktitle}{\emph{Proceedings of the 2021 International Conference on Management of Data}}. \bibinfo{pages}{2614--2627}.
\newblock


\bibitem[\protect\citeauthoryear{Wang, Zhang, Zhang, Lin, and Cheema}{Wang et~al\mbox{.}}{2015}]%
        {wang2015optimal}
\bibfield{author}{\bibinfo{person}{Xiaoyang Wang}, \bibinfo{person}{Ying Zhang}, \bibinfo{person}{Wenjie Zhang}, \bibinfo{person}{Xuemin Lin}, {and} \bibinfo{person}{Muhammad~Aamir Cheema}.} \bibinfo{year}{2015}\natexlab{}.
\newblock \showarticletitle{Optimal spatial dominance: an effective search of nearest neighbor candidates}. In \bibinfo{booktitle}{\emph{Proceedings of the 2015 ACM SIGMOD International Conference on Management of Data}}. \bibinfo{pages}{923--938}.
\newblock


\bibitem[\protect\citeauthoryear{Wei, Wu, Wang, Lou, Zhan, Li, and Cai}{Wei et~al\mbox{.}}{2020}]%
        {adb-v}
\bibfield{author}{\bibinfo{person}{Chuangxian Wei}, \bibinfo{person}{Bin Wu}, \bibinfo{person}{Sheng Wang}, \bibinfo{person}{Renjie Lou}, \bibinfo{person}{Chaoqun Zhan}, \bibinfo{person}{Feifei Li}, {and} \bibinfo{person}{Yuanzhe Cai}.} \bibinfo{year}{2020}\natexlab{}.
\newblock \showarticletitle{AnalyticDB-V: a hybrid analytical engine towards query fusion for structured and unstructured data}.
\newblock \bibinfo{journal}{\emph{Proceedings of the VLDB Endowment}} \bibinfo{volume}{13}, \bibinfo{number}{12} (\bibinfo{year}{2020}), \bibinfo{pages}{3152--3165}.
\newblock


\bibitem[\protect\citeauthoryear{Wieschollek, Wang, Sorkine-Hornung, and Lensch}{Wieschollek et~al\mbox{.}}{2016}]%
        {wieschollek2016efficient}
\bibfield{author}{\bibinfo{person}{Patrick Wieschollek}, \bibinfo{person}{Oliver Wang}, \bibinfo{person}{Alexander Sorkine-Hornung}, {and} \bibinfo{person}{Hendrik Lensch}.} \bibinfo{year}{2016}\natexlab{}.
\newblock \showarticletitle{Efficient large-scale approximate nearest neighbor search on the gpu}. In \bibinfo{booktitle}{\emph{Proceedings of the IEEE Conference on Computer Vision and Pattern Recognition}}. \bibinfo{pages}{2027--2035}.
\newblock


\bibitem[\protect\citeauthoryear{Wu, Jin, and Zhang}{Wu et~al\mbox{.}}{2014}]%
        {wu2014fast}
\bibfield{author}{\bibinfo{person}{Yubao Wu}, \bibinfo{person}{Ruoming Jin}, {and} \bibinfo{person}{Xiang Zhang}.} \bibinfo{year}{2014}\natexlab{}.
\newblock \showarticletitle{Fast and unified local search for random walk based k-nearest-neighbor query in large graphs}. In \bibinfo{booktitle}{\emph{Proceedings of the 2014 ACM SIGMOD international conference on Management of Data}}. \bibinfo{pages}{1139--1150}.
\newblock


\bibitem[\protect\citeauthoryear{Xu, Alon, and Neubig}{Xu et~al\mbox{.}}{2023a}]%
        {xu2023nearest}
\bibfield{author}{\bibinfo{person}{Frank~F Xu}, \bibinfo{person}{Uri Alon}, {and} \bibinfo{person}{Graham Neubig}.} \bibinfo{year}{2023}\natexlab{a}.
\newblock \showarticletitle{Why do Nearest Neighbor Language Models Work?}
\newblock \bibinfo{journal}{\emph{arXiv preprint arXiv:2301.02828}} (\bibinfo{year}{2023}).
\newblock


\bibitem[\protect\citeauthoryear{Xu, Liang, Li, Xu, Chen, Zhang, Li, Yang, Yang, Yang, et~al\mbox{.}}{Xu et~al\mbox{.}}{2023b}]%
        {xu2023spfresh}
\bibfield{author}{\bibinfo{person}{Yuming Xu}, \bibinfo{person}{Hengyu Liang}, \bibinfo{person}{Jin Li}, \bibinfo{person}{Shuotao Xu}, \bibinfo{person}{Qi Chen}, \bibinfo{person}{Qianxi Zhang}, \bibinfo{person}{Cheng Li}, \bibinfo{person}{Ziyue Yang}, \bibinfo{person}{Fan Yang}, \bibinfo{person}{Yuqing Yang}, {et~al\mbox{.}}} \bibinfo{year}{2023}\natexlab{b}.
\newblock \showarticletitle{SPFresh: Incremental In-Place Update for Billion-Scale Vector Search}. In \bibinfo{booktitle}{\emph{Proceedings of the 29th Symposium on Operating Systems Principles}}. \bibinfo{pages}{545--561}.
\newblock


\bibitem[\protect\citeauthoryear{Yang, Cheema, Lin, and Wang}{Yang et~al\mbox{.}}{2015}]%
        {yang2015reverse}
\bibfield{author}{\bibinfo{person}{Shiyu Yang}, \bibinfo{person}{Muhammad~Aamir Cheema}, \bibinfo{person}{Xuemin Lin}, {and} \bibinfo{person}{Wei Wang}.} \bibinfo{year}{2015}\natexlab{}.
\newblock \showarticletitle{Reverse k nearest neighbors query processing: experiments and analysis}.
\newblock \bibinfo{journal}{\emph{Proceedings of the VLDB Endowment}} \bibinfo{volume}{8}, \bibinfo{number}{5} (\bibinfo{year}{2015}), \bibinfo{pages}{605--616}.
\newblock


\bibitem[\protect\citeauthoryear{Yang, Li, Fang, and Wei}{Yang et~al\mbox{.}}{2020}]%
        {yang2020pase}
\bibfield{author}{\bibinfo{person}{Wen Yang}, \bibinfo{person}{Tao Li}, \bibinfo{person}{Gai Fang}, {and} \bibinfo{person}{Hong Wei}.} \bibinfo{year}{2020}\natexlab{}.
\newblock \showarticletitle{Pase: Postgresql ultra-high-dimensional approximate nearest neighbor search extension}. In \bibinfo{booktitle}{\emph{Proceedings of the 2020 ACM SIGMOD international conference on management of data}}. \bibinfo{pages}{2241--2253}.
\newblock


\bibitem[\protect\citeauthoryear{Yogatama, de~Masson~d’Autume, and Kong}{Yogatama et~al\mbox{.}}{2021}]%
        {yogatama2021adaptive}
\bibfield{author}{\bibinfo{person}{Dani Yogatama}, \bibinfo{person}{Cyprien de Masson~d’Autume}, {and} \bibinfo{person}{Lingpeng Kong}.} \bibinfo{year}{2021}\natexlab{}.
\newblock \showarticletitle{Adaptive semiparametric language models}.
\newblock \bibinfo{journal}{\emph{Transactions of the Association for Computational Linguistics}}  \bibinfo{volume}{9} (\bibinfo{year}{2021}), \bibinfo{pages}{362--373}.
\newblock


\bibitem[\protect\citeauthoryear{Yu, Jeong, Kim, Kim, and Chun}{Yu et~al\mbox{.}}{2022}]%
        {yu2022orca}
\bibfield{author}{\bibinfo{person}{Gyeong-In Yu}, \bibinfo{person}{Joo~Seong Jeong}, \bibinfo{person}{Geon-Woo Kim}, \bibinfo{person}{Soojeong Kim}, {and} \bibinfo{person}{Byung-Gon Chun}.} \bibinfo{year}{2022}\natexlab{}.
\newblock \showarticletitle{Orca: A distributed serving system for $\{$Transformer-Based$\}$ generative models}. In \bibinfo{booktitle}{\emph{16th USENIX Symposium on Operating Systems Design and Implementation (OSDI 22)}}. \bibinfo{pages}{521--538}.
\newblock


\bibitem[\protect\citeauthoryear{Zeng, Zhu, Liu, Zhang, Dai, Zhou, Li, Ning, Xie, Yang, et~al\mbox{.}}{Zeng et~al\mbox{.}}{2023}]%
        {zeng2023df}
\bibfield{author}{\bibinfo{person}{Shulin Zeng}, \bibinfo{person}{Zhenhua Zhu}, \bibinfo{person}{Jun Liu}, \bibinfo{person}{Haoyu Zhang}, \bibinfo{person}{Guohao Dai}, \bibinfo{person}{Zixuan Zhou}, \bibinfo{person}{Shuangchen Li}, \bibinfo{person}{Xuefei Ning}, \bibinfo{person}{Yuan Xie}, \bibinfo{person}{Huazhong Yang}, {et~al\mbox{.}}} \bibinfo{year}{2023}\natexlab{}.
\newblock \showarticletitle{DF-GAS: a Distributed FPGA-as-a-Service Architecture towards Billion-Scale Graph-based Approximate Nearest Neighbor Search}.
\newblock  (\bibinfo{year}{2023}).
\newblock


\bibitem[\protect\citeauthoryear{Zhang, Khoram, and Li}{Zhang et~al\mbox{.}}{2018}]%
        {zhang2018efficient}
\bibfield{author}{\bibinfo{person}{Jialiang Zhang}, \bibinfo{person}{Soroosh Khoram}, {and} \bibinfo{person}{Jing Li}.} \bibinfo{year}{2018}\natexlab{}.
\newblock \showarticletitle{Efficient large-scale approximate nearest neighbor search on OpenCL FPGA}. In \bibinfo{booktitle}{\emph{Proceedings of the IEEE Conference on Computer Vision and Pattern Recognition}}. \bibinfo{pages}{4924--4932}.
\newblock


\bibitem[\protect\citeauthoryear{Zhang, Xu, Chen, Sui, Xie, Cai, Chen, He, Yang, Yang, et~al\mbox{.}}{Zhang et~al\mbox{.}}{2023}]%
        {zhang2023vbase}
\bibfield{author}{\bibinfo{person}{Qianxi Zhang}, \bibinfo{person}{Shuotao Xu}, \bibinfo{person}{Qi Chen}, \bibinfo{person}{Guoxin Sui}, \bibinfo{person}{Jiadong Xie}, \bibinfo{person}{Zhizhen Cai}, \bibinfo{person}{Yaoqi Chen}, \bibinfo{person}{Yinxuan He}, \bibinfo{person}{Yuqing Yang}, \bibinfo{person}{Fan Yang}, {et~al\mbox{.}}} \bibinfo{year}{2023}\natexlab{}.
\newblock \showarticletitle{$\{$VBASE$\}$: Unifying Online Vector Similarity Search and Relational Queries via Relaxed Monotonicity}. In \bibinfo{booktitle}{\emph{17th USENIX Symposium on Operating Systems Design and Implementation (OSDI 23)}}. \bibinfo{pages}{377--395}.
\newblock


\bibitem[\protect\citeauthoryear{Zhao, Tian, Huang, Zheng, and Zhou}{Zhao et~al\mbox{.}}{2023}]%
        {zhao2023towards}
\bibfield{author}{\bibinfo{person}{Xi Zhao}, \bibinfo{person}{Yao Tian}, \bibinfo{person}{Kai Huang}, \bibinfo{person}{Bolong Zheng}, {and} \bibinfo{person}{Xiaofang Zhou}.} \bibinfo{year}{2023}\natexlab{}.
\newblock \showarticletitle{Towards efficient index construction and approximate nearest neighbor search in high-dimensional spaces}.
\newblock \bibinfo{journal}{\emph{Proceedings of the VLDB Endowment}} \bibinfo{volume}{16}, \bibinfo{number}{8} (\bibinfo{year}{2023}), \bibinfo{pages}{1979--1991}.
\newblock


\bibitem[\protect\citeauthoryear{Zheng, Guo, Tung, and Wu}{Zheng et~al\mbox{.}}{2016}]%
        {zheng2016lazylsh}
\bibfield{author}{\bibinfo{person}{Yuxin Zheng}, \bibinfo{person}{Qi Guo}, \bibinfo{person}{Anthony~KH Tung}, {and} \bibinfo{person}{Sai Wu}.} \bibinfo{year}{2016}\natexlab{}.
\newblock \showarticletitle{Lazylsh: Approximate nearest neighbor search for multiple distance functions with a single index}. In \bibinfo{booktitle}{\emph{Proceedings of the 2016 International Conference on Management of Data}}. \bibinfo{pages}{2023--2037}.
\newblock


\bibitem[\protect\citeauthoryear{Zhu, Yang, Wang, and Lee}{Zhu et~al\mbox{.}}{2016}]%
        {zhu2016range}
\bibfield{author}{\bibinfo{person}{Huaijie Zhu}, \bibinfo{person}{Xiaochun Yang}, \bibinfo{person}{Bin Wang}, {and} \bibinfo{person}{Wang-Chien Lee}.} \bibinfo{year}{2016}\natexlab{}.
\newblock \showarticletitle{Range-based obstructed nearest neighbor queries}. In \bibinfo{booktitle}{\emph{Proceedings of the 2016 International Conference on Management of Data}}. \bibinfo{pages}{2053--2068}.
\newblock


\bibitem[\protect\citeauthoryear{Zuo and Deng}{Zuo and Deng}{2023}]%
        {zuo2023arkgraph}
\bibfield{author}{\bibinfo{person}{Chaoji Zuo} {and} \bibinfo{person}{Dong Deng}.} \bibinfo{year}{2023}\natexlab{}.
\newblock \showarticletitle{ARKGraph: All-Range Approximate K-Nearest-Neighbor Graph}.
\newblock \bibinfo{journal}{\emph{Proceedings of the VLDB Endowment}} \bibinfo{volume}{16}, \bibinfo{number}{10} (\bibinfo{year}{2023}), \bibinfo{pages}{2645--2658}.
\newblock


\end{thebibliography}

\end{document}